\documentclass{article}

\usepackage[preprint]{aistats2026}
\usepackage{natbib}

\usepackage[utf8]{inputenc} % allow utf-8 input
\usepackage[T1]{fontenc}    % use 8-bit T1 fonts
\usepackage{hyperref}       % hyperlinks
\usepackage{url}            % simple URL typesetting
\usepackage{booktabs}       % professional-quality tables
\usepackage{amsfonts}       % blackboard math symbols
\usepackage{amsmath}
\usepackage{nicefrac}       % compact symbols for 1/2, etc.
\usepackage{microtype}      % microtypography
\usepackage{xcolor}         % colors
\usepackage{amssymb}
\usepackage{bm}
\usepackage{multirow} 
\usepackage{float}
\usepackage{graphicx}
\usepackage{subcaption}
\usepackage{tcolorbox}
\tcbset{colback=yellow!20!white, colframe=red!75!black}

\usepackage{mathtools}
\usepackage{amsthm}
\usepackage{algorithm}
\usepackage{algorithmic}
\newtheoremstyle{definitionstyle}
{10pt} % Space above
{10pt} % Space below
{} % Body font
{} % Indent amount
{\bfseries} % Theorem head font
{.} % Punctuation after theorem head
{ } % Space after theorem head
{} % Theorem head spec

\usepackage{xcolor}
\usepackage{hyperref}
\hypersetup{
    colorlinks,
    linkcolor={red!50!black},
    citecolor={blue!50!black},
    urlcolor={blue!80!black}
}

\DeclareMathOperator{\tr}{tr}

\DeclareMathOperator*{\argmin}{arg\,min}

\newcommand{\RR}{{\mathbb{R}}}

\newcommand{\PP}{{\mathbb{P}}}
\newcommand{\EE}{{\mathbb{E}}}
\newcommand{\NN}{{\mathcal{N}}}

\newcommand{\A}{\mathbf{A}}
\newcommand{\B}{\mathbf{B}}

\newcommand{\V}{\mathbf{V}}
\newcommand{\U}{\mathbf{U}}

\newcommand{\X}{\mathbf{X}}
\newcommand{\Z}{\mathbf{Z}}
\newcommand{\Q}{\mathbf{Q}}
\newcommand{\Y}{\mathbf{Y}}

\newcommand{\ba}{\mathbf{a}}
\newcommand{\bb}{\mathbf{b}}

\newcommand{\x}{\mathbf{x}}
\newcommand{\y}{\mathbf{y}}
\newcommand{\z}{\mathbf{z}}
\newcommand{\rr}{\mathbf{r}}

\newcommand{\w}{\mathbf{w}}
\newcommand{\bv}{\mathbf{v}}
\newcommand{\bu}{\mathbf{u}}
\newcommand{\zo}{\mathbf{0}}
\newcommand{\one}{\mathbf{1}}
\newcommand{\I}{\mathbf{I}}

\newcommand{\bbeta}{\boldsymbol{\beta}}

\newcommand{\cd}{{ \xrightarrow{d} }}
\newcommand{\asto}{{ \xrightarrow{a.s.} }}

\definecolor{RED}{rgb}{0.7,0,0}
\definecolor{BLUE}{rgb}{0,0,0.69}
\definecolor{GREEN}{rgb}{0,0.6,0}
\definecolor{PURPLE}{rgb}{0.69,0,0.8}
\definecolor{ORANGE}{RGB}{255,103,0}
\definecolor{BROWN}{RGB}{100,20,45}

\theoremstyle{definitionstyle}
\newtheorem{definition}{Definition}

\usepackage[capitalize,noabbrev]{cleveref}
\usepackage{tcolorbox}
\usepackage{graphicx}
\usepackage{listings}
\usepackage{xcolor}
\usepackage{subcaption}
\usepackage{booktabs}
\usepackage{float}    % for [H] placement if desired
\usepackage{sidecap}  % Allows side captions

\lstdefinestyle{greenstyle}{
	language=Python,
	backgroundcolor=\color{green!10},      % light green background
	basicstyle=\ttfamily\small\color{green!50!black}, % dark green text
	frame=single,
	rulecolor=\color{green!80!black},       % dark green frame
	breaklines=true,
	numbers=left,                          % optional: show line numbers
	numberstyle=\tiny\color{green!50!black},
	stepnumber=1,
	numbersep=10pt,
}

\theoremstyle{plain}
\newtheorem{theorem}{Theorem}[section]
\newtheorem{proposition}[theorem]{Proposition}
\newtheorem{lemma}[theorem]{Lemma}

\theoremstyle{definition}
\newtheorem{assumption}[theorem]{Assumption}
\theoremstyle{remark}
\newtheorem{remark}[theorem]{Remark}

\begin{document}
\twocolumn[

\aistatstitle{A Linear Approach to Data Poisoning}

\aistatsauthor{ D. G. M. Flynn \And Diego Granziol }

\aistatsaddress{ Mathematical Institute, Oxford \And  Mathematical Institute, Oxford} ]

\begin{abstract}
Backdoor and data-poisoning attacks can flip predictions with tiny training corruptions, yet a sharp theory linking poisoning strength, overparameterization, and regularization is lacking. We analyze ridge least squares with an unpenalized intercept in the high-dimensional regime \(p,n\to\infty\), \(p/n\to c\). Targeted poisoning is modelled by shifting a \(\theta\)-fraction of one class by a direction \(\mathbf{v}\) and relabelling. Using resolvent techniques and deterministic equivalents from random matrix theory, we derive closed-form limits for the poisoned score explicit in the model parameters. The formulas yield scaling laws, recover the interpolation threshold as \(c\to1\) in the ridgeless limit, and show that the weights align with the poisoning direction. Synthetic experiments match theory across sweeps of the parameters and MNIST backdoor tests show qualitatively consistent trends. The results provide a tractable framework for quantifying poisoning in linear models.
\end{abstract}

% \section{Random Matrix Analysis of Linear Data Poisoning}

% In this section, we provide a random matrix theory framework to characterize the effect of targeted data poisoning on high-dimensional linear least-square models. We place ourselves under an analytically tractable model in which both the data and the poisoning mechanism are random, allowing for explicit spectral analysis in the large-dimensional regime.

\section{Introduction}
\label{sec:introduction}

Backdoor and data-poisoning attacks can subvert modern ML systems with small, targeted changes to the training data. Empirically, even tiny poisoned fractions imprint triggers that flip predictions at test time while leaving standard accuracy intact, across supervised and contrastive pretraining regimes \citep{gu2017badnets,turner_label-consistent_2019,shafahi_poison_2018,carlini_poisoning_2022,xu2024shadowcast}. Despite extensive empirical evidence and many attack variants, sharp theory quantifying \emph{how} poisoning strength, overparameterization, and regularization interact to control attack efficacy remains limited. 

Recent research has extended the study of data poisoning and backdoor attacks into the domain of large language models (LLMs), demonstrating that vulnerabilities persist even at scale. Instruction-tuned LLMs can be manipulated with only a small fraction of poisoned examples, leading to persistent triggers that reliably alter outputs without degrading overall utility \citep{zhou_learning_2025}. Beyond classical poisoning, methods such as direct model editing enable attackers to implant backdoors efficiently into pretrained weights \citep{li_badedit_2024}. Benchmarks confirm that LLMs present multiple attack surfaces, including fine-tuning pipelines, prompt injection, and system-level instructions, making them especially susceptible to both prompt-based triggers \citep{yao_poisonprompt_2023} and stealthy manipulation of instruction hierarchies \citep{zhang_instruction_2024,zhou_aspire_2024}. Recent analyses further indicate that poisoning in LLMs interacts with data contamination and emergent behavior in complex ways, highlighting that backdoor robustness must be considered a critical property of foundation models \citep{he_multi-faceted_2025,ge_when_2024}.

\paragraph{AI ethics and governance motivation.}
Backdoor and data-poisoning risks are recognized as first-class threats in modern AI governance. The EU AI Act explicitly lists data poisoning as a cyber-risk for “high-risk’’ systems, mandating robustness, transparency, and lifecycle security \citep{EUAIActArt15,EUAIActRecital76}. The NIST AI Risk Management Framework and AML taxonomy similarly highlight poisoning as a cross-cutting vulnerability requiring measurable safeguards and incident response \citep{NIST_RMF,NIST_GenAI_Profile,NIST_AML_Taxonomy}. Operational frameworks such as MITRE ATLAS and OWASP’s Top-10 for LLM/GenAI catalog poisoning and backdoor tactics as key adversarial classes \citep{MITRE_ATLAS,OWASP_LLM_2025,OWASP_LLM_2023}. Accountability artifacts—including Datasheets for Datasets, Model Cards, and AI FactSheets—stress provenance and risk documentation, for which solution-shift diagnostics (e.g., weight projections, curvature analysis) offer auditable evidence \citep{Gebru2021Datasheets,Mitchell2019ModelCards,Arnold2019Factsheets}. Beyond compliance, malicious-use analyses underscore the societal stakes of covert model manipulation \citep{Brundage2018MaliciousUse}, while sustainability perspectives emphasize efficient defenses that enable targeted remediation without full retraining \citep{Schwartz2020GreenAI}. Together, these considerations motivate principled methods that measure and explain how solutions move under poisoning, aligning technical robustness with governance and accountability.

\paragraph{AI theory motivation.}
Early work on general poisoning without backdoor triggers established worst-case guarantees for ERM with outlier-removal defences \citep{Steinhardt2017Certified} and showed, in the regression setting, how bi-level optimization attacks can shift learned solutions while trimmed estimators offer provable robustness \citep{Jagielski2018RegressionPoisoning}. A distinct line of work exploits the structure of backdoors: poisoned examples form mixtures with shifted means, producing identifiable spectral signatures that align with the trigger direction \citep{Tran2018SpectralSignatures}, with later refinements using robust statistics to tolerate higher contamination \citep{Hayase2021Spectre} and certified detectors that guarantee detectability when shifts exceed a threshold \citep{Xiang2023CBD}. Others provide robustness certificates against triggers, for example through randomized smoothing \citep{Wang2020SmoothingBackdoor} or federated learning schemes with clipping and smoothing \citep{Xie2021CRFL}, while mechanistic analyses formalize when backdoors are learnable, tying vulnerability to excess capacity \citep{ManojBlum2021ExcessCapacity}, orthogonality of gradients and decision regions \citep{Zhang2024Orthogonality}, and adaptability directions in classical learners \citep{Xian2023Adaptability}; at the extreme, cryptographic constructions show that undetectable backdoors exist in principle \citep{Goldwasser2022UndetectableBackdoors}. Whereas these works primarily establish detection criteria, certification bounds, or conditions for learnability, our contribution is complementary: we provide closed-form, high-dimensional limit laws for poisoned ridge least squares, giving explicit scaling in $(\theta,|\bv|,\lambda,c)$ and showing estimator alignment with the trigger direction, thereby turning qualitative vulnerability into quantitative \emph{solution deltas} that can serve as auditable, governance-ready measures of poisoning efficacy.

\paragraph{RMT theory motivation.}
Random matrix theory yields sharp asymptotics for ridge regression in the proportional high-dimensional regime. Classical results give deterministic equivalents for spectra and risk, leading to closed-form formulas for prediction error and optimal regularization \citep{Dobriban2018Highdimridge,Hastie2019Surprises,couillet_random_2022,Mei2019Generalization}. These works position RMT as a lens on bias–variance trade-offs, double descent, and kernel generalization, but they have not been developed from a security or poisoning perspective. In particular, while spectral methods underpin many empirical defenses against backdoors, a full RMT framework for quantifying adversarial vulnerabilities or solution shifts under poisoning remains essentially absent.

% \paragraph{Motivation.}
% Understanding the dynamics of poisoning is not merely of academic interest: backdoor risks are treated as first-class threats in modern AI ethics and governance frameworks. The EU AI Act, for example, explicitly lists data poisoning as a relevant cyber-risk for ``high-risk'' AI systems, requiring robustness and transparency throughout the lifecycle. For both safety and certification purposes, we need principled methods that measure and explain how the learned solution shifts under adversarial perturbations of the training set, rather than relying solely on heuristic defenses.

% \paragraph{Related work.}
% Most prior work has focused on developing and demonstrating increasingly powerful poisoning and backdoor attacks \citep{gu2017badnets,turner_label-consistent_2019,shafahi_poison_2018,carlini_poisoning_2022} or on heuristic defenses that attempt to detect and filter poisoned data \citep{tran2018spectral,perry2022doresnet}. While these contributions highlight the empirical severity of the threat, the theoretical side has remained underdeveloped. Existing analyses largely characterize poisoning in linear models or under restrictive assumptions \citep{steinhardt2017certified,schwarzschild2021just}, leaving open the question of how attacks scale with modern overparameterized architectures, implicit bias from optimization, and regularization.

\section{Contributions and Results in brief}
Our work develops a tractable random matrix framework for quantifying the impact of targeted backdoor poisoning on high-dimensional ridge least squares. The main findings and contributions are as follows. 

\begin{tcolorbox}[colback=blue!5!white,colframe=blue!75!black,title=Takeaway I: Bigger is Better]
% \textbf{TPoisoning robustness.}
For the same amount of data, larger models are more robust to a fixed fraction of data poisoning.
\end{tcolorbox}
\begin{tcolorbox}[colback=blue!5!white,colframe=blue!75!black,title=Takeaway II: Regularisation does not help]
% \textbf{TPoisoning robustness.}
% To first order, there is no effect on data poisoning efficacy as we alter regularisation.
The effect of regularisation is small, as the contributions from the mean and variance are in opposite directions.
\end{tcolorbox}
% % 
% \begin{tcolorbox}[colback=blue!5!white,colframe=blue!75!black,title=Take away for Practitioners]
% % \textbf{TPoisoning robustness.}
% In the proportional high-dimensional limit, increasing model and data size makes robustness provably higher for a fixed fixed fraction of poisoned data, compared to smaller models. 
% \end{tcolorbox}

\paragraph{Framework and scope.}
We analyze ridge least squares with an unpenalized intercept in the high-dimensional regime $p,n\to\infty$ with $p/n\to c\in(0,\infty)$. We assume a random data and labels model, in order to focus on the effect of adding the poisoning. The theoretical work assumes that labels are independent of the data, however we include MNIST experiments that go beyond this setting.

Poisoning is targeted and structured: a $\theta$-fraction of one class is shifted by a direction $\bv$ and relabelled. This Gaussian, rank-one contamination model provides analytic clarity and yields testable predictions. Section~\ref{sec:model} details the assumptions, Section~\ref{sec:main-result} gives the analytic results, Section~\ref{sec:proof-sketch} outlines the proof strategy, and Section~\ref{sec:discussion} reports synthetic and image-domain experiments.

\begin{figure}[htbp]
    \centering
    \begin{subfigure}[b]{0.48\columnwidth}
        \centering
        \includegraphics[width=\textwidth]{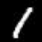}
        \caption{Unpoisoned}
        \label{fig:unpoisoned}
    \end{subfigure}
    \hfill
    \begin{subfigure}[b]{0.48\columnwidth}
        \centering
        \includegraphics[width=\textwidth]{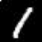}
        \caption{Poisoned}
        \label{fig:poisoned}
    \end{subfigure}
    \caption{An example of the poisoning mechanism, a small perturbation is made in the top left of the image. The left image is correctly labelled as "$1$", and the right image has its label changed to "$0$" in the poisoned dataset.}
    \label{fig:poison-comparison}
\end{figure}

\paragraph{Model motivation}
We wish to develop the simplest model that is able to capture the essence of data poisoning. We will then test the predictions from our model against real data to verify that we have captured true behaviour. To this end, we consider a signal free model, formed of Gaussian data $\x$, which is mapped to a randomly chosen label $y \in \{-1, +1\}$. We then perform the ``poisoning'', by adding a spike (a vector $\bv$) to our data $\x$ and changing the label, a setting considered empirically for example in \cite{gu2017badnets} Section 4. In an image setting this corresponds to \ref{fig:poison-comparison}.

Our empirical claim is that the shift caused in the high dimensional regressor from this spike is largely independent of the underlying data distribution, and so whether this is a signal free Gaussian, or the highly non-Gaussian anisotropic MNIST dataset, we see similar behaviour Figure (\ref{fig:mu-comparison}). By studying in depth the former, we are then able to match trends and get insight into the latter.

We note additionally that we model the poisoning direction to be fixed rather than trained. We take some fixed $\bv$ and examine the behaviour based on this (and indeed we see that this depends only on $\| \bv \|$ in our model). This is a distinct phenomenon to that of adversarial attacks, which has received large amounts of attention in the literature beginning with \cite{szegedy_intriguing_2014}. In an adversarial attack, the model is taken as given and the dataset is assumed to be ``clean'', and the attack is to find a small perturbation of the input that results in a shift in the prediction.

\textbf{Closed-form limit law.} We prove that the poisoned predictor admits an asymptotic Gaussian distribution, 
\[
\hat\bbeta^{\top}(\x_{0}+\bv)\ \cd\ \mathcal{N}(\mu,\sigma^{2}),
\]
with explicit expressions for the mean shift $\mu(\theta,\|\bv\|,\lambda,c)$ and variance $\sigma^{2}(\theta,\|\bv\|,\lambda,c)$. These formulas yield a direct efficacy proxy
\[
\eta = 1 - \Phi(-\mu/\sigma)
\]
for backdoor success, where $\Phi$ is the CDF of $\mathcal{N}(0, 1)$.  

\textbf{Alignment with the trigger.} The estimator $\hat\bbeta$ concentrates onto the poisoning direction $\bv$: the alignment coefficient grows with the poisoning rate $\theta$ and diminishes with stronger regularization $\lambda$, making explicit how ridge penalization counteracts trigger imprinting.  

\textbf{Scaling laws and interpolation effects.} The theory isolates the separate roles of poisoning fraction, trigger norm, regularization, and aspect ratio. In particular, in the ridgeless limit the variance term reproduces the interpolation threshold blow-up as $c \to 1$, and we also see for small $\theta$ that the poison efficacy has a linear effect in $\theta$.

\textbf{Validation.} Synthetic experiments confirm the closed-form predictions across wide sweeps of $(c,\lambda,\|\bv\|,\theta)$, while MNIST backdoor experiments reproduce the predicted scaling trends for the mean shift $\mu$ and show qualitatively consistent though noisier variance behaviour. These findings support the use of the linear RMT theory as a heuristic lens for understanding poisoning in more complex models.

\section{Model and assumptions}
\label{sec:model}
We take a clean dataset, with isotropic Gaussian vectors, and we assume the labels are unstructured. Since we are including the centring in our model (via the bias) to remove the mean, this assumption amounts to discounting the covariance structure of our dataset.
\begin{assumption}[Clean data]
\label{ass:clean-data}
Let \(\X = [\x_1,\dots,\x_n] \in \mathbb{R}^{p\times n}\) be the data matrix and \(\y=(y_1,\dots,y_n)\in\{-1,+1\}^n\) the labels. Assume
\[
\PP(y_i=1)=\PP(y_i=-1)=\tfrac12 \quad \text{independently over } i,
\]
and that
\(
\x_i \sim 
\mathcal{N}(0,\I_p)
\)
All samples \((\x_i,y_i)\) are independent across \(i\).
\end{assumption}

We analyse the case where the poisoning amounts to adding a small feature perturbation to elements of one class, while changing the label to the other class in a binary setting. This follows the backdoor poisoning setup originally proposed in \cite{gu2017badnets}.

\begin{assumption}[Poisoning mechanism]
\label{ass:poison}
Fix \(\theta\in[0,1]\) and a direction \(\bv\in\mathbb{R}^p\) with \(\|\bv\|=O(1)\). Define an indicator vector \(\bu\in\{0,1\}^n\) with
\[
\PP(u_i=1\,|\,y_i)=
\begin{cases}
\theta, & y_i=-1,\\
0, & y_i=+1,
\end{cases}
\qquad\text{independently over } i.
\]
For \(u_i=1\) the attacker shifts the feature by \(\bv\) and flips its label to \(+1\); otherwise the sample is unchanged. Equivalently,
\[
\x_i^{\text{pois}}=\x_i+\bv\,u_i, 
\qquad 
y_i^{\text{pois}}=y_i+2u_i,
\]
or, in matrix form,
\[
\X^{\text{pois}}=\X+\bv\bu^\top, 
\qquad 
\y^{\text{pois}}=\y+2\bu.
\]
Conditionally, this yields
\[
\x_i^{\text{pois}} \sim 
\begin{cases}
\mathcal{N}(0,\I_p), &  u_i=0.\\[2pt]
\mathcal{N}(\bv,\I_p), & u_i=1.
\end{cases}
\]
\end{assumption}

Finally, we assume that we are in a high-dimensional regime, where $p$ the dimension of our data and $n$ the number of data points are comparable and large. This regime is common in random matrix analysis of machine learning, and is able to much more accurately capture behaviour than the classical limit where we fix $p$ and take $n \rightarrow \infty$ \cite{couillet2022random, vershynin_high-dimensional_2018, wainwright_high-dimensional_2019}. We take $p, n \rightarrow \infty$ in our analytic results, and we verify the results experimentally for large fixed $p, n$.

\begin{assumption}[High-dimensional regime]
\label{ass:high-dim}
We let $n,p \to \infty$ with $p/n \to c \in (0,\infty)$. 
All parameters $(\theta,\lambda,c)$ are fixed as $n,p\to\infty$ and we assume that $\|\bv \| = O(1)$.
\end{assumption}

\paragraph{Model.}
We use ridge least squares with an \emph{unpenalized} intercept $b_0$:
\begin{align}
\mathcal{L}(\bbeta,b_0)
&=\frac1n\sum_{i=1}^n\big(\bbeta^{\top}\x_i^\text{pois}+b_0-y_i^\text{pois}\big)^2+\lambda\|\bbeta\|^2 \\
(\hat\bbeta, \hat b_0) &= \argmin_{(\bbeta ,  b_0) \in \RR^p \times \RR} \mathcal{L}(\bbeta,b_0;\X^\text{pois},\y^\text{pois},\lambda)
\end{align}
Let $\bar\x=\EE[\frac1n\X^\text{pois}\one ]$, $\bar w= \EE [\frac1n\one^\top\y^\text{pois} ]$, and define centred variables
$\widetilde\X=\X^\text{pois}-\bar\x\one^\top$, $\widetilde\w=\y^\text{pois}-\bar w\one$.
The estimator equals
\begin{equation}
\label{eqn:beta_explicit_form}
\hat\bbeta=\frac1n\Big(\frac1n\widetilde\X\widetilde\X^\top+\lambda\I_p\Big)^{-1}\widetilde\X\,\widetilde\w,
\qquad
\hat b_0=\bar w-\hat\bbeta^\top\bar\x .
\end{equation}
In particular, under assumptions \ref{ass:clean-data} and \ref{ass:poison} we have
\(
\bar\x=\tfrac{\theta}{2}\bv,
\ \bar w=\theta
\),
and with $\rr:=\bu-\tfrac{\theta}{2}\one$,
\(
\widetilde\X=\X+\bv\,\rr^\top,\ 
\widetilde\w=\y+2\rr
\).

\section{Main result}
\label{sec:main-result}

\begin{figure*}[t]
  \centering
  {\small\bfseries Synthetic}\\[0.25em]
  \begin{subfigure}{0.32\textwidth}
    \includegraphics[width=\linewidth]{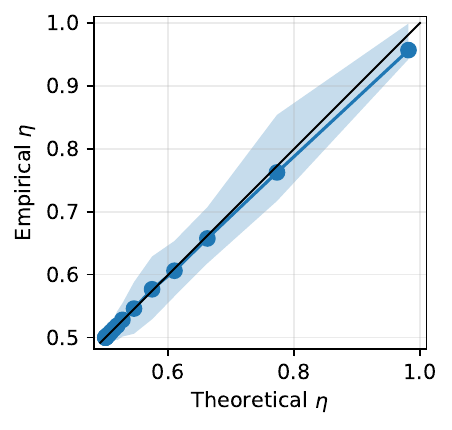}
  \end{subfigure}\hfill
  \begin{subfigure}{0.32\textwidth}
    \includegraphics[width=\linewidth]{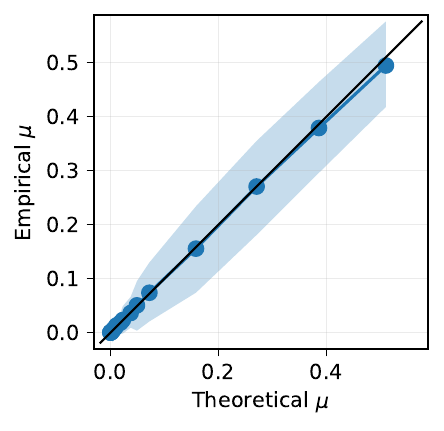}
  \end{subfigure}\hfill
  \begin{subfigure}{0.32\textwidth}
    \includegraphics[width=\linewidth]{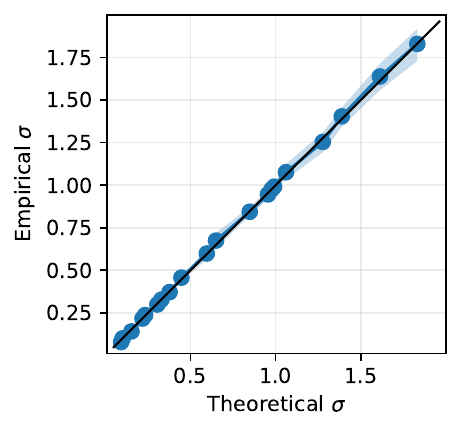}
  \end{subfigure}
  \caption{We plot empirical vs theoretical values for the derived $\eta$, $\mu$ and $\sigma$ under the synthetic data model. We perform a parameter sweep across $c$, $\lambda$, $\| \bv \|$ and $\theta$, plotting the theoretical parameter against the empirical for each case. The circular marks are the averages over 100 independent trials, and the shaded region is the interquartile range. }
  \label{fig:synthetic-emp-vs-theory}
\end{figure*}

\begin{figure*}[t]
% \centering
%   \begin{minipage}{0.75\textwidth} % 75% width
  \centering
{\small\bfseries Synthetic}\\[0.25em]
  \begin{subfigure}{0.48\textwidth}
    \includegraphics[width=\linewidth]{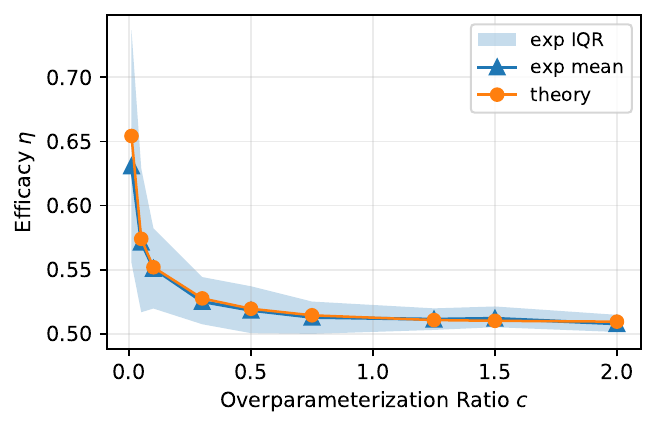}
    % \caption{A}
  \end{subfigure}\hfill
  \begin{subfigure}{0.48\textwidth}
    \includegraphics[width=\linewidth]{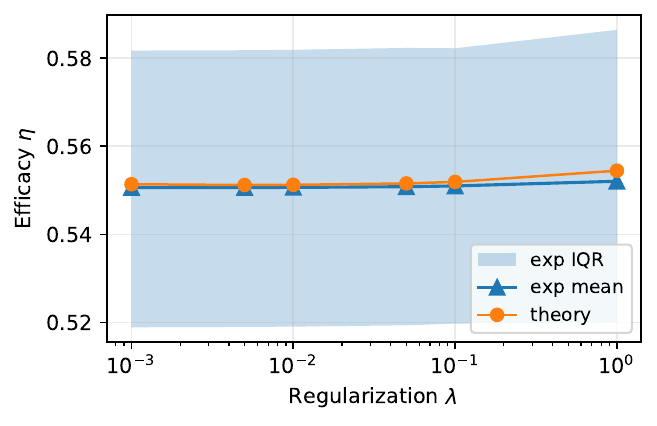}
    % \caption{B}
  \end{subfigure}

  \vspace{0.6em}

  \begin{subfigure}{0.48\textwidth}
    \includegraphics[width=\linewidth]{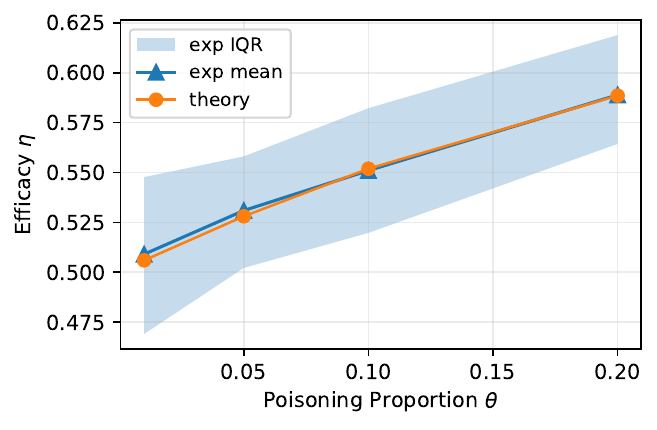}
    % \caption{C}
  \end{subfigure}\hfill
  \begin{subfigure}{0.48\textwidth}
    \includegraphics[width=\linewidth]{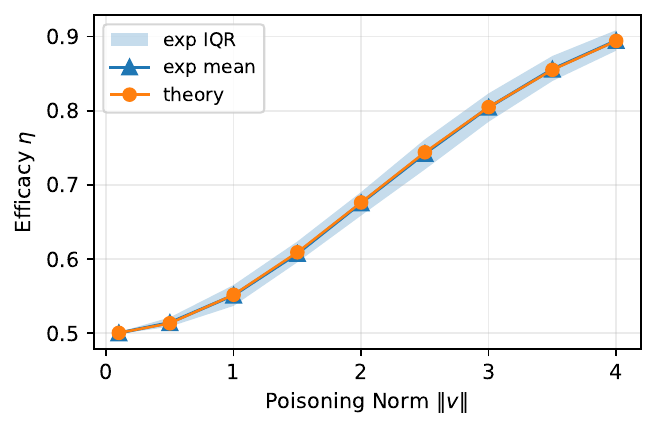}
    % \caption{D}
  \end{subfigure}
  % \end{minipage}

  \caption{Plot of the poisoning efficacy $\eta$ for the binary classifier on synthetic data, as the parameters vary. We use the fixed values $c=0.1$, $\lambda=0.1$, $\| \bv \| = 1$ and $\theta=0.1$, and then vary each parameter in turn. The circles are the theoretical results, and the triangles are the experimental results, with the interquartile range shaded.}
  \label{fig:synthetic-eta}
\end{figure*}

\begin{figure*}[t]

  \centering
  % Top row: theory
{\small\bfseries Synthetic}\\[0.25em]
  \begin{subfigure}{0.32\textwidth}\includegraphics[width=\linewidth]{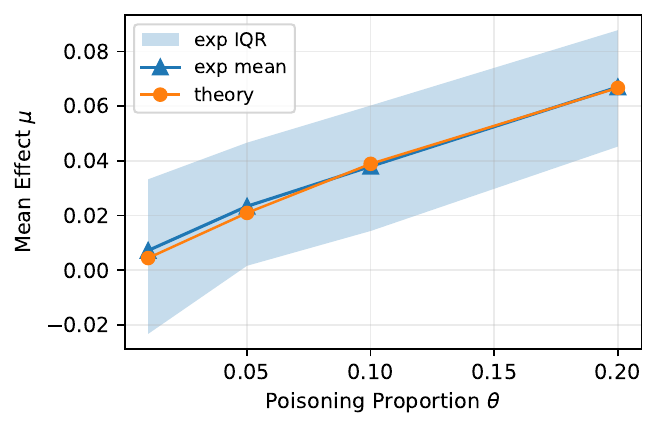}\end{subfigure}\hfill
  \begin{subfigure}{0.32\textwidth}\includegraphics[width=\linewidth]{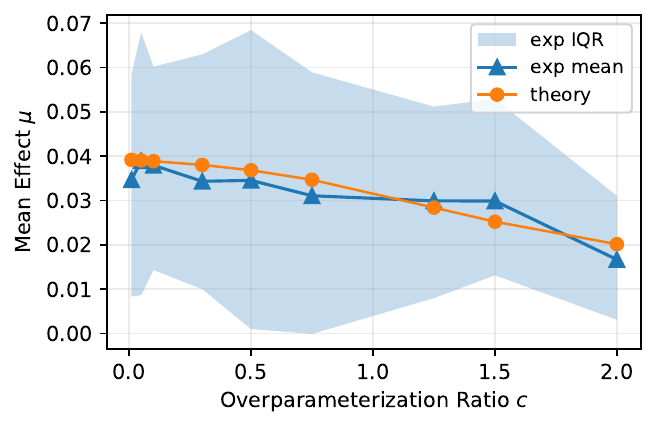}\end{subfigure}\hfill
  \begin{subfigure}{0.32\textwidth}\includegraphics[width=\linewidth]{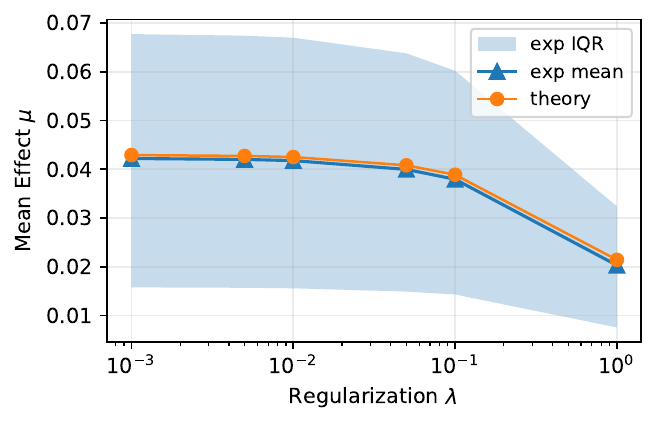}\end{subfigure}

  \vspace{0.6em}

  % Bottom row: practice
  {\small\bfseries MNIST}\\[0.25em]
  \begin{subfigure}{0.32\textwidth}\includegraphics[width=\linewidth]{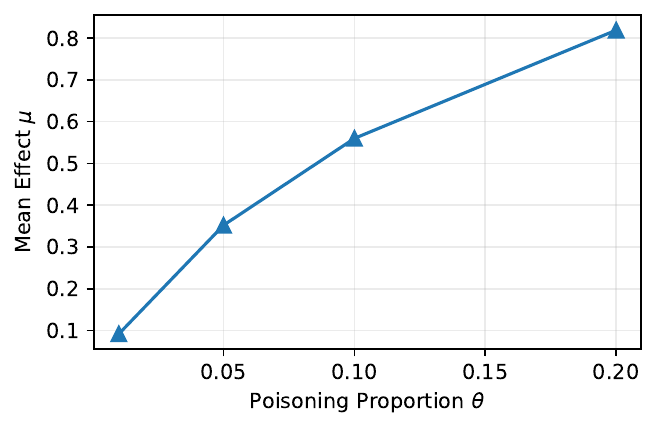}\end{subfigure}\hfill
  \begin{subfigure}{0.32\textwidth}\includegraphics[width=\linewidth]{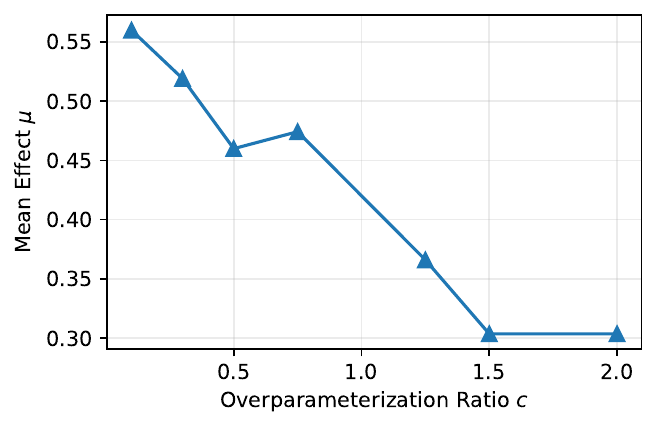}\end{subfigure}\hfill
  \begin{subfigure}{0.32\textwidth}\includegraphics[width=\linewidth]{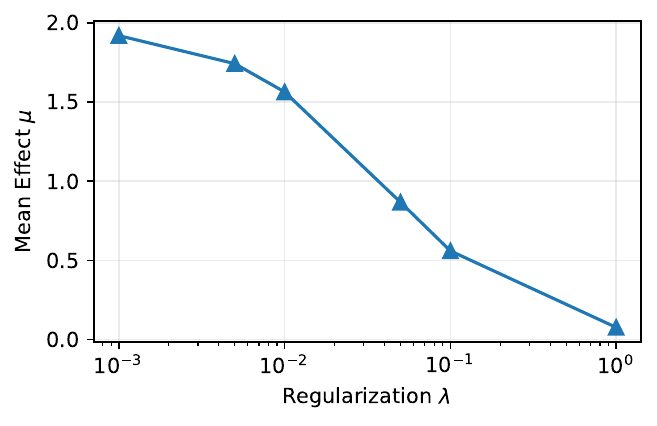}\end{subfigure}

  \caption{Top: empirical and theoretical results for the $\mu$ in the synthetic data model. We fix the values $c=0.1$, $\|\bv\| = 1$, $\theta = 0.1$, $\lambda = 0.1$, and then vary $c$, $\theta$ and $\lambda$. We fix $p=500$ and vary $n$ to get the required values of $c$. We plot the experimental mean and IQR across 100 independent samples. Bottom: We perform the poisoning and classification on MNIST. We use the same values of $c$, $\| \bv \|$, $\theta$ and $\lambda$, and vary the number of data points included to control $c$. We do a binary classification on the digits "$0$" and "$1$".}
  \label{fig:mu-comparison}
\end{figure*}

\begin{figure*}[t]

  \centering
  % Top row: theory
{\small\bfseries Synthetic}\\[0.25em]
  \begin{subfigure}{0.32\textwidth}\includegraphics[width=\linewidth]{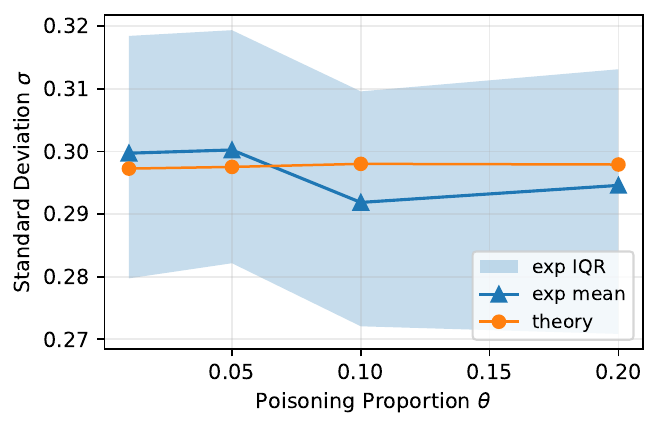}\end{subfigure}\hfill
  \begin{subfigure}{0.32\textwidth}\includegraphics[width=\linewidth]{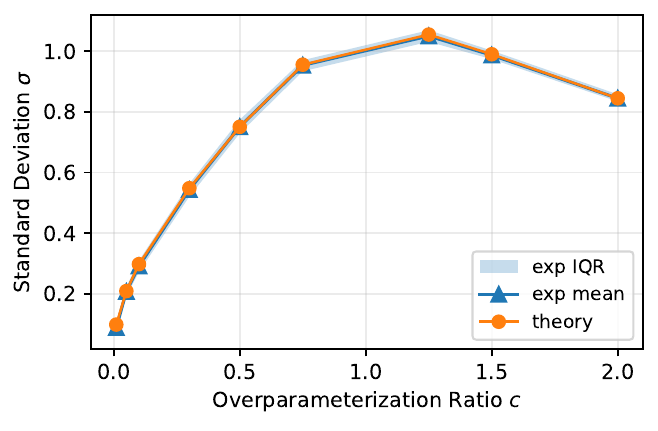}\end{subfigure}\hfill
  \begin{subfigure}{0.32\textwidth}\includegraphics[width=\linewidth]{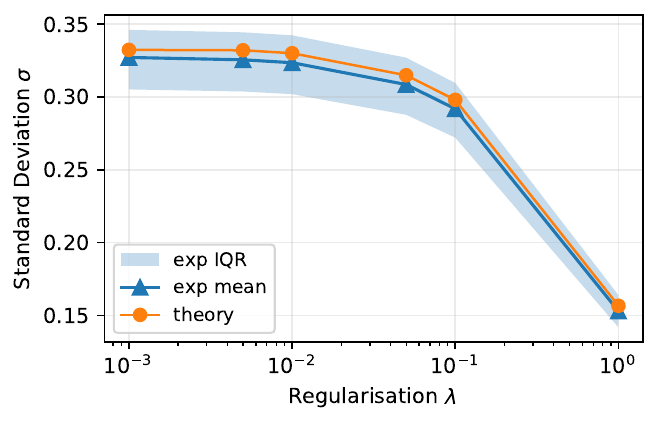}\end{subfigure}

  \vspace{0.6em}

  % Bottom row: practice
  {\small\bfseries MNIST}\\[0.25em]
  \begin{subfigure}{0.32\textwidth}\includegraphics[width=\linewidth]{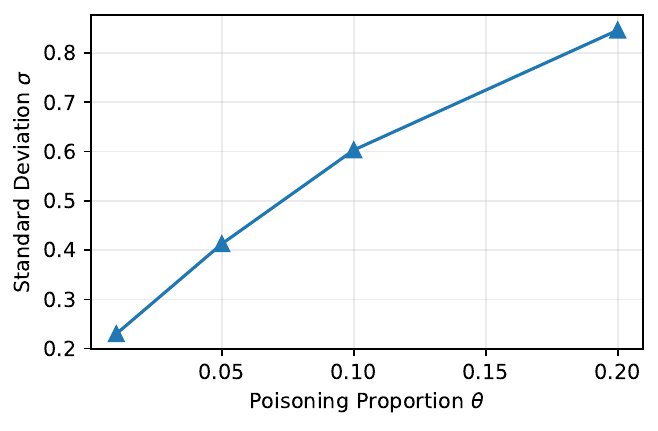}\end{subfigure}\hfill
  \begin{subfigure}{0.32\textwidth}\includegraphics[width=\linewidth]{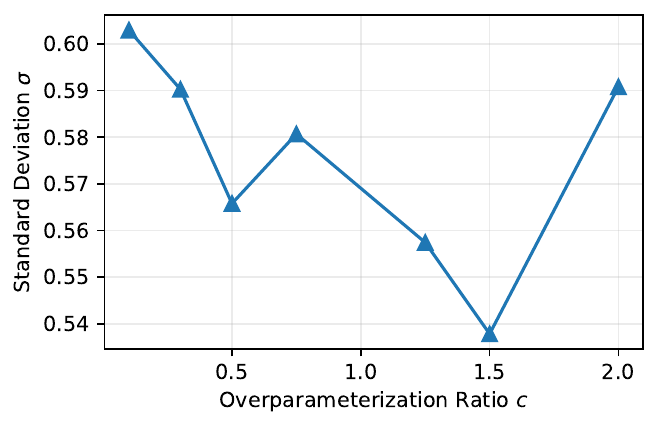}\end{subfigure}\hfill
  \begin{subfigure}{0.32\textwidth}\includegraphics[width=\linewidth]{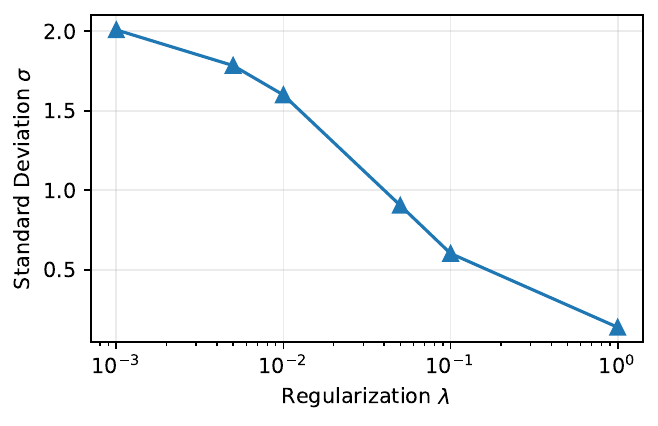}\end{subfigure}

  \caption{
  Comparison for $\sigma$. The setup of the experiments ia the same as in Figure \ref{fig:mu-comparison}.
  % Top: empirical and theoretical results for the $\mu$ in the synthetic data model. We fix the values $c=0.1$, $\|\bv\| = 1$, $\theta = 0.1$ $\lambda = 0.1$, and then vary $c$, $\theta$ and $\lambda$. We fix $p=500$ and vary $n$ to get the required values of $c$. We plot the experimental mean and IQR across 100 independent samples. Bottom: We perform the poisoning and regression on MNIST. We use the same values of $c$, $\| \bv \|$, $\theta$ and $\lambda$, and vary the number of data points included to control $c$. We do a binary classification on the digits "$0$" and "$1$".
  }
  \label{fig:sigma-comparison}
\end{figure*}

We evaluate the behaviour of the poisoned model on a new independent poisoned test sample $\x_0 +  \bv$, where $\x_0 \sim \mathcal{N}(0, \I_p)$ is independent of the training set.

\begin{theorem}[Asymptotic distribution under poisoning]
Under assumptions \ref{ass:clean-data}-\ref{ass:high-dim}, as $n, p \to \infty$ with $p/n \to c \in (0, \infty)$, the scalar output of the poisoned regressor satisfies
\begin{equation}
\hat{\bbeta}^{\top}(\x_0 + \bv) \stackrel{D}{\to} \mathcal{N}(\mu, \sigma^2),
\end{equation}
where
\begin{equation}
\label{eqn:mu-final}
\mu = \frac{ \theta (1 - \theta) m(-\lambda)  \| \bv \|^2 }{(1 + cm(-\lambda))(2+\| \bv \|^2 \theta(1 - \tfrac{\theta}{2})(1 - \lambda m(-\lambda)))},
\end{equation}
\begin{equation}\label{eq:sigma2-final}
\begin{aligned}
\sigma^2
&=\Big(\tilde m(-\lambda)-\lambda \tilde m'(-\lambda)\Big)
\bigg[
(1-\theta^2)\\
& +\left(\frac{1+c^{-1}\tau^2}{\big(1+c^{-1}\tau^2\,(1-\lambda \tilde m(-\lambda))\big)^2}-1\right)
\frac{\theta(1-\theta)^2}{\,2-\theta\,} \bigg],
\end{aligned}
\end{equation}

with \(\tau^2=\|\bv\|^2\,\frac{\theta}{2}\!\left(1-\frac{\theta}{2}\right)\).

and $m(z)$ is the Stieltjes transform of the Marčenko–Pastur distribution with parameter $c$, given by
\begin{equation}
m(z) = \frac{1- c - z - \sqrt{(1-c-z)^2 - 4cz}}{2cz},
\end{equation}
and
\begin{equation}
\tilde{m}(z) = cm(z) - \frac{1-c}{z}.
\end{equation}

\end{theorem}

\begin{remark}[Limit for vanishing regularization]
When $\lambda \to 0$ and $c < 1$, (\ref{eqn:mu-final})–(\ref{eq:sigma2-final}) reduce to
\begin{align}
\mu_0 &= \frac{\| \bv \|^2\theta(1-\theta)}{2 + \|\bv\|^2 \theta(1 - \tfrac{\theta}{2})}, \\
\sigma_0^2&=\frac{c}{1-c}\left[
(1-\theta^2)
+\left(\frac{1+c^{-1}\tau^2}{(1+\tau^2)^2}-1\right)
\frac{\theta(1-\theta)^2}{\,2-\theta\,}
\right].
\end{align}
\end{remark}

\begin{remark}[Poisoning efficacy]
For a binary classifier with decision threshold 0, the poisoning efficacy is
\begin{equation}
\eta_{\text{pois}} = 1 - \Phi\left(-\frac{\mu}{\sigma}\right),
\end{equation}
where $\Phi$ is the CDF of $\mathcal{N}(0, 1)$. For small $\theta$, the model predicts $\eta_{\text{pois}} \approx 0.5 + O(\theta)$, consistent with empirical observations.
\end{remark}
% \begin{remark}[Dependence on the parameters]
%     We note that $\bmu$ is an increasing function of $\theta$ for small $\theta$ and 
% \end{remark}

\begin{proposition}[Alignment with poisoning vector]
Under assumptions \ref{ass:clean-data}-\ref{ass:high-dim}
the estimator $\hat{\bbeta}$ aligns asymptotically with $\bv$:
\begin{equation}
\begin{gathered}
\hat{\bbeta}^{\top}\ba \stackrel{\text{a.s.}}{\to} C \, \bv^{\top}\ba, \\
C = \frac{ \theta (1 - \theta) m(-\lambda)}{(1 + cm(-\lambda))(2+\| \bv \|^2 \theta(1 - \tfrac{\theta}{2})(1 - \lambda m(-\lambda)))},
\end{gathered}
\end{equation}
for any deterministic $\ba \in \mathbb{R}^p$. Thus $C$ increases with $\theta$ and decreases with $\lambda$.
\end{proposition}

\section{Proof sketch}
\label{sec:proof-sketch}

We outline the argument leading to the Gaussian limit and the closed forms for $\mu$ and $\sigma^2$. Full proofs, including concentration and leave-one-out estimates, are deferred to the Appendix.

\paragraph{Methods}

 The overarching idea is to use the Woodbury Lemma, an algebraic matrix identity:

\begin{lemma}[Woodbury Formula (circa 1950)]
	For $\mathbf{A} \in \mathbb{R}^{p \times p}$, $\mathbf{U}, \mathbf{V} \in \mathbb{R}^{p \times d}$, such that both $\mathbf{A}$ and $\mathbf{A} + \mathbf{U}\mathbf{V}^\top$ are invertible, we have
	\begin{align*}
	(\mathbf{A} + \mathbf{U}\mathbf{V}^\top)^{-1} = \mathbf{A}^{-1} - \mathbf{A}^{-1}\mathbf{U}(\mathbf{I}_d + \mathbf{V}^\top\mathbf{A}^{-1}\mathbf{U})^{-1}\mathbf{V}^\top\mathbf{A}^{-1}
	\end{align*}
\end{lemma}
 Crucially here, if $p$ is large and $k$ is finite (in our case we use this for $k=3$), then the inner matrix here is a small $k\times k$ inverse, and hence we do not have any issues interchanging matrix inversion and limits. This is fundamentally what allows us to control the behaviour of random matrices when they have a low-rank perturbation. This is used together with the fact that the unpoisoned Gaussian (Wishart) matrix has a well understood resolvent.

 Specifically, we have that if $\X = [\x_1, \ldots \x_n] \in \RR^{p \times n}$ is an isotropic Gaussian matrix, in the sense that $\x_i \sim \mathcal{N}(0, \I_p)$. Then defining
 \[\Q_0(z) := \left(\frac{1}{n}\X \X^\top - z \I_p\right)^{-1},\] we have that \[\Q_0(z) \longleftrightarrow m(z) \I_p. \] By "$\longleftrightarrow$" we mean becomes deterministically equivalent to - a notion we make precise in the appendix, where here $m(z)$ is the Stieltjes transform of the Mar\v{c}enko-Pastur distribution - the connection to Stieltjes transforms comes in here as $\tr \Q_0$ is the Stieltjes transform of the empirical distribution of the eigenvalues of $\X$. Informally however, this tells us that when we are taking linear functionals of the matrix $\Q_0$, then we can simply replace it with $m(z) \I_p$.

     % To put these together, we use the fact that the poisoned matrix is a rank $1$ perturbation of our Gaussian matrix $\widetilde \X = \X + \bv \rr^\top$, and hence we can expand the gram matrix, and get a rank $3$ perturbation of our Wishart matrix $\X \X^\top$
     % \[
     % 	\frac1n \widetilde \X \widetilde \X^\top = \frac1n \X \X^\top + \frac{1}{n}  \bv \rr^\top \X^\top + \frac{1}{n}  \X \rr \bv^\top + \frac{1}{n}\bv \bv ^\top
     % \]
\paragraph{Step 1: Centering and linearization.}
Recall the ridge estimator with unpenalized intercept from~\eqref{eqn:beta_explicit_form}:
\[
\begin{aligned}
\hat\bbeta
&=\frac1n\Big(\tfrac1n\widetilde\X\widetilde\X^\top+\lambda\I_p\Big)^{-1}
 \widetilde\X\,\widetilde\w,\\
\hat b_0
&=\bar w-\hat\bbeta^\top\bar\x .
\end{aligned}
\]
With $\rr=\bu-\tfrac{\theta}{2}\one$ we have
\[
\begin{aligned}
\widetilde\X&=\X+\bv \rr^\top,\\
\widetilde\w&=\y+2\rr ,
\end{aligned}
\]
and the LLNs
\[
\begin{aligned}
\frac1n\|\rr\|^2 &\asto s:=\frac{\theta}{2}\Big(1-\frac{\theta}{2}\Big),\\
\frac1n\,\y^\top\rr &\asto -\frac{\theta}{2},\\
\frac1n\,\rr^\top\widetilde\w &\asto \frac{\theta(1-\theta)}{2}.
\end{aligned}
\]
Let $z=-\lambda$, $\Q_1(z)=(\tfrac1n\widetilde\X\widetilde\X^\top-z\I_p)^{-1}$ and
$\tilde\Q_1(z)=(\tfrac1n\widetilde\X^\top\widetilde\X-z\I_n)^{-1}$. Then
\[
\hat\bbeta=\frac1n \Q_1(-\lambda)\,\widetilde\X\,\widetilde\w.
\]

\paragraph{Step 2: Rank-one spike and resolvent deterministic equivalents.}
Write the spike direction $\ba:=\bv/\|\bv\|$ and set
\[
\tau^2:=\|\bv\|^2\,s=\|\bv\|^2\,\frac{\theta}{2}\Big(1-\frac{\theta}{2}\Big).
\]
The perturbation $\widetilde\X=\X+\bv\rr^\top$ is rank one on each side of $\X$.

By the Woodbury Lemma, and the Deterministic equivalent of the Wishart Matrix, the resolvents admit deterministic equivalents
that depend only on $m(z)$, $\tilde m(z)$, and $\tau^2$ (see Appendix Lemmas A.1-A.2 following \citet{couillet2022random}):
\[
\begin{aligned}
\Q_1(z) &\Longleftrightarrow \bar\Q_1(z) \\
\bar \Q_1(z)&=  m(z)\!\Bigg(
\I_p
- \frac{\tau^2(1+z m(z))}{1+\tau^2(1+z m(z))}\ba\ba^\top
\Bigg),\\
\tilde\Q_1(z) &\Longleftrightarrow \tilde m(z)\!\Bigg(
\I_n
- \frac{c^{-1}\tau^2(1+z\tilde m(z))}{1+c^{-1}\tau^2(1+z\tilde m(z))}\bar\bb\,\bar\bb^\top
\Bigg),
\end{aligned}
\]
where $\bar\bb:=\rr/\|\rr\|$. For the squared resolvent one has
\[
\begin{aligned}
\tilde\Q_1(z)^2
&\Longleftrightarrow \tilde m'(z)\I_n
+\Big(T(z)-\tilde m'(z)\Big)\bar\bb\,\bar\bb^\top,\\
T(z)
&=\frac{(c^{-1}\tau^2+1)\tilde m'(z)-c^{-1}\tau^2\tilde m(z)^2}
{\big(1+c^{-1}\tau^2(1+z\tilde m(z))\big)^2}.
\end{aligned}
\]
\paragraph{Step 3: Mean of the poisoned score.}
We study the scalar $\hat\bbeta^\top\bv$. Expanding $\widetilde\X=\X+\bv\rr^\top$,
\[
\begin{aligned}
\hat\bbeta
&=\frac1n \Q_1(-\lambda)\big(\X\y + 2\X\rr \\
&\qquad\qquad\quad + \bv\,\rr^\top\y + 2\bv\,\rr^\top\rr\big).
\end{aligned}
\]
By independence and centering, $\EE[\X\y]=\EE[\X\rr]=\zo$, so the contribution comes from the $\bv$-term. A standard leave-one-out argument (Sherman–Morrison on $\Q_1$) yields the decoupling
\[
\begin{aligned}
\frac1n \Q_1(-\lambda)\,\bv\,\rr^\top\widetilde\w
&=\frac{1}{1+c\,m(-\lambda)}\,\bar\Q_1(-\lambda)\,\bv\;\cdot\;\frac{1}{n}\,\rr^\top\widetilde\w \\
&\quad + o_{\|\cdot\|}(1),
\end{aligned}
\]
where $m$ is the Marčenko–Pastur Stieltjes transform. Using $\frac1n\,\rr^\top\widetilde\w\asto \frac{\theta(1-\theta)}{2}$ and $\bar\Q_1(-\lambda)\,\bv=m(-\lambda)\bv/(1+\tau^2(1-\lambda m(-\lambda)))$, we obtain
\[
\begin{aligned}
\EE[\hat\bbeta^\top\bv]
&=\frac{m(-\lambda)}{1+c\,m(-\lambda)}\cdot
\frac{\tfrac{1}{2} \|\bv\|^2\theta(1-\theta)}{1+\tau^2\big(1-\lambda m(-\lambda)\big)}
+o(1).
\end{aligned}
\]
Rewriting $1+\tau^2(1-\lambda m)=\frac{1}{2}\big(2+\|\bv\|^2\theta(1-\tfrac{\theta}{2})(1-\lambda m)\big)$ gives the announced
\[
\mu=\frac{ \theta (1 - \theta) m(-\lambda) \|\bv\|^2}{
(1 + c\,m(-\lambda))
\big(2+\|\bv\|^2 \theta(1 - \tfrac{\theta}{2})(1 - \lambda m(-\lambda))\big)}.
\]
A Gaussian concentration (e.g. quadratic-forms close-to-trace and a Burkholder inequality) shows $\hat\bbeta^\top\bv-\EE[\hat\bbeta^\top\bv]\to 0$; see Appendix.

\paragraph{Step 4: Variance via $\|\hat\bbeta\|^2$.}
For a fresh test point $\x_0\sim\NN(0,\I_p)$ independent of the training set, the only asymptotic randomness in $\hat\bbeta^\top(\x_0+\bv)$ comes from $\x_0$ because $\hat\bbeta$ concentrates. Conditionally on $\hat\bbeta$, $\hat\bbeta^\top\x_0\sim\NN(0,\|\hat\bbeta\|^2)$, hence the asymptotic variance is $\sigma^2=\lim \|\hat\bbeta\|^2$.

Using the identity
\[
\frac{1}{n}\,\widetilde\X^\top \Q_1(z)^2 \widetilde\X
= z\,\tilde\Q_1(z)^2+\tilde\Q_1(z),
\]
we obtain
\[
\|\hat\bbeta\|^2
=\frac{1}{n}\,\widetilde\w^\top\!\big(z\,\tilde\Q_1(z)^2+\tilde\Q_1(z)\big)\widetilde\w
\Big|_{z=-\lambda}.
\]
Insert the Deterministic Equivalent from Step~2 and the LLNs 
\[
\begin{aligned}
\frac{1}{n}\|\widetilde\w\|^2 &\asto 1-\theta^2,\\
\frac{1}{n}\big(\widetilde\w^\top\bar\bb\big)^2
&\asto \frac{\theta(1-\theta)^2}{\,2-\theta\,}.
\end{aligned}
\]
Let $B(z)=1+c^{-1}\tau^2(1+z\tilde m(z))$ and $S(z)=\tilde m(z)+z\tilde m'(z)$. A short algebraic manipulation (Appendix) shows
\[
\begin{aligned}
z\big(T(z)-\tilde m'(z)\big)
&-\tilde m(z)\!\left(1-\frac{1}{B(z)}\right) \\
&= S(z)\!\left(\frac{c^{-1}\tau^2+1}{B(z)^2}-1\right).
\end{aligned}
\]
Therefore,
\[
\begin{aligned}
&\|\hat\bbeta\|^2 \rightarrow
S(-\lambda)\!\bigg[
(1-\theta^2) \\
&+\left(\frac{1+c^{-1}\tau^2}{\big(1+c^{-1}\tau^2(1-\lambda \tilde m(-\lambda))\big)^2}-1\right)
\frac{\theta(1-\theta)^2}{\,2-\theta\,}
\bigg],
\end{aligned}
\]
which is exactly~\eqref{eq:sigma2-final} since $S(-\lambda)=\tilde m(-\lambda)-\lambda \tilde m'(-\lambda)$.

\paragraph{Step 5: Gaussian limit for the poisoned score.}
Conditionally on the training set, $\hat\bbeta^\top\x_0\sim\NN(0,\|\hat\bbeta\|^2)$ and $\hat\bbeta^\top\bv\to\mu$. Since $\|\hat\bbeta\|^2\to\sigma^2$ in probability and $\x_0$ is independent, we get
\[
\hat\bbeta^\top(\x_0+\bv)\ \cd\ \NN(\mu,\sigma^2).
\]

\paragraph{Step 6: Alignment.}
For any deterministic $\ba\in\RR^p$,
\[
\EE[\hat\bbeta^\top \ba]
= \frac{\tfrac{1}{2}\theta(1-\theta)}{1+c\,m(-\lambda)}\,\ba^\top\bar\Q_1(-\lambda)\bv + o(1)
\ \to\ C\,\bv^\top\ba,
\]
with
\[
C=\frac{ \theta (1 - \theta) m(-\lambda)}{(1 + c\,m(-\lambda))\big(2+\|\bv\|^2 \theta(1 - \tfrac{\theta}{2})(1 - \lambda m(-\lambda))\big)}.
\]
Thus $\hat\bbeta$ asymptotically lies in the span of $\bv$ and the alignment increases with~$\theta$ and decreases with~$\lambda$.

\paragraph{Step 7: Checks and limits.}
(i) No poisoning ($\theta=0$) gives $\mu=0$ and $\sigma^2=\tilde m(-\lambda)-\lambda\tilde m'(-\lambda)$ (clean ridge). 
(ii) As $\lambda\to0$ with $c<1$, use $\tilde m(-\lambda) - \lambda\tilde m'(-\lambda) \rightarrow \frac{c}{1-c}$ and $ 1- \lambda \tilde m(-\lambda) \rightarrow c$ to recover the unregularized limits reported in the remarks. 
(iii) For small $\theta$, $\mu/\sigma=O(\theta)$, matching the linear growth of attack efficacy.

\medskip
Throughout we repeatedly use (a) Woodbury/Sherman–Morrison to isolate single-sample and spike effects; (b) deterministic equivalents for spiked resolvents; (c) quadratic-form close-to-trace lemmas and Gaussian concentration to replace random denominators by their limits; and (d) leave-one-out resolvents to decouple dependencies between $\widetilde\z_i$ and $\Q_1$. Precise statements and bounds are in the Appendix, following the methodology of \citet{couillet2022random}.

\section{Discussion and Experiments}
\label{sec:discussion}

We perform experiments on the MNIST dataset \cite{deng_mnist_2012}, as well as the synthetic dataset generated procedurally by the setup of Assumptions \ref{ass:clean-data}, \ref{ass:poison}: we generate a Gaussian matrix with assigned labels and then apply the poisoning to a proportion $\theta$ of the data. We use a new random seed over each retry to produce the interquartile range shown on the figures. For both the synthetic and MNIST datasets, we perform the experiments using the explicit Equation \ref{eqn:beta_explicit_form}, computing the matrix computations on an Nvidia 3080Ti GPU. We perform a parameter sweep over a range of $(\lambda, c, \| \bv \|, \theta)$.

\begin{table}[h]
\centering
\begin{tabular}{ll}
\hline
Parameter & Sweep Values \\
\hline
$c$      & 0.1, 0.3, 0.5, 0.75, 1.25, 1.5, 2.0 \\
$\lambda$ & 0.001, 0.005, 0.01, 0.05, 0.1, 1.0 \\
$\theta$  & 0.01, 0.05, 0.1, 0.2 \\
$\|\bv\|$   & 0.1, 0.5, 1, 1.5, 2, 2.5, 3, 3.5, 4 \\
\hline
\end{tabular}
\caption{Parameter grid used for the sweep.}
\end{table}
We choose the $\bv$ direction as shown in Figure \ref{fig:poison-comparison}. We fix the dimension $p=500$ for each synthetic plot, and vary the dataset size $n$ accordingly to get required values of $c$. Similarly we vary $n$ for the MNIST case to get the values of $c=784/n$. We then compute $\mu$ via $\hat \bbeta^\top \bv$ and $\sigma^2$ as $\hat \bbeta^\top \hat \bbeta$. We restrict to the binary case on MNIST, to the digits 0 and 1, labelled with $\{-1, +1\}$.

On the synthetic dataset, we find a strong verification of the theory, as shown in Figures \ref{fig:synthetic-emp-vs-theory} and \ref{fig:synthetic-eta}. On each plot of the synthetic data we plot the empirically observed values in blue against the theoretical values from \eqref{eqn:mu-final}–\eqref{eq:sigma2-final} in orange. 
% \diegoinsert{what does this mean? the trends in $\mu$ and $\sigma$ match as in how they change with what? Then if they don't match we could of course recallibrate each scale to match, what would this mean?}. 
Increasing poison rate $\theta$ and trigger strength $\|\bv\|$ increases the mean shift $\mu$ and also increases the variance $\sigma$ via $\tau^2$, yielding higher attack success $\eta=1-\Phi(-\mu/\sigma)$. In the synthetic dataset we see stronger regularization $\lambda$ reduces $\mu$ and $\sigma$ (with a stronger effect on $\sigma$) resulting in an increase of $\eta$ with $\lambda$. Larger aspect ratio $c=p/n$ decreases $\mu$ and has a more varying effect on $\sigma$ resulting in a general decrease in $\eta$ as $c$ increases.

On the MNIST dataset, we find good qualitative agreement between the theory and the practice on the mean (first-order effect) of the poisoning $\mu$ (Fig.~\ref{fig:mu-comparison}). We see that the mean increases as $\theta$ increases, decreases with $\lambda$ and decreases as $c$ increases. In particular this means on a larger dataset of the same dimension, hence with $c=p/n$ smaller, then  the average effect of poisoning is larger,\textit{ even while keeping the proportion the same}.
% \diegoinsert{what does this mean exactly? that as we increase the dataset size poisoning becomes more effective? at the same proportion?}

For the second order effect $\sigma$, we see a divergence of behaviour between the synthetic data model and the MNIST data, as seen in Figure \ref{fig:sigma-comparison}. A plausible cause is violation of the isotropy assumption: correlated features and class structure increase output variance depending on the alignment between $\bv$ and high-variance directions.

\section{Limitations}

Our analysis has several limitations. First, we assume isotropic features and a fixed poisoning (backdoor) direction, which greatly simplifies the random matrix analysis but does not capture more realistic structured covariances or adaptive attack strategies. Second, the clean labels are taken to be randomized and independent of the features, so the model does not include an underlying predictive signal; this allows exact characterization of the poisoning effect but limits direct applicability to typical supervised learning problems where $\x$ carries information about $y$. Third, we focus on linear ridge regression with an unpenalized intercept, thereby excluding both more complex linear estimators (e.g., with feature-dependent regularization) and the nonlinear models commonly used in practice. Finally, all results are derived in an asymptotic high-dimensional regime, so finite-sample deviations may be non-negligible on real datasets. Extending the framework to structured covariance (e.g., anisotropic features), informative labels, and non-quadratic losses or more general empirical risk minimization problems is an important direction for future work. Extending to anisotropic features would be a feasible extension using an anisotropic deterministic equivalent lemma in place of the Wishart deterministic equivalent. (e.g. Theorem 2.6 from \cite{couillet_random_2022}.)

A natural extension of our framework is to models with hierarchical or crossed random effects, which are widely used in large-scale recommendation and feed-ranking systems. In such settings one typically augments a linear or logistic predictor with group-specific random offsets (e.g., user, item, or session effects), estimated via scalable mixed-model machinery \citep[e.g.][]{ghosh_scalable_2021, bellio_consistent_2025}. From a poisoning perspective, this structure creates both challenges and opportunities: an attacker can concentrate corruptions on a specific group or cluster, effectively steering its random effect, and the efficacy of such a targeted attack should depend on the corresponding variance component—large random-effect variances permit substantial group-level deviations from the global trend, while strong shrinkage would attenuate the impact of group-confined poisoning. At the same time, the explicit estimation of random effects and variance components offers potential detection signals, for example by monitoring anomalous group-level effects or abrupt changes in variance estimates over time. Developing an RMT-style analysis of poisoning in these hierarchical models, and formalizing detection criteria based on random-effect diagnostics, is an interesting direction that we leave for future work.

\section{Conclusion}
\label{sec:conclusion}

We have developed a random matrix framework that quantifies the impact of structured backdoor poisoning in high-dimensional ridge least squares. By deriving closed-form asymptotic laws for the poisoned score, we identified how poisoning rate, trigger strength, regularization, and overparameterization jointly govern attack efficacy. Experiments on synthetic data and MNIST confirm the theory’s predictions, supporting its use as a heuristic lens for more complex models. Looking ahead, extending this analysis to correlated features, nonlinear learners, and modern representation models may provide the same combination of analytic clarity and governance-ready diagnostics that we demonstrated in the linear setting.

\section{Acknowledgements}
The authors would also acknowledge support from His Majesty’s Government in the development of
this research. DF is funded by the Charles Coulson Scholarship.

\newpage
\bibliographystyle{iclr2025_conference}
\bibliography{bib,zotero,iclr2025_conference}

@article{xu2024shadowcast,
	title={Shadowcast: Stealthy data poisoning attacks against vision-language models},
	author={Xu, Yuancheng and Yao, Jiarui and Shu, Manli and Sun, Yanchao and Wu, Zichu and Yu, Ning and Goldstein, Tom and Huang, Furong},
	journal={arXiv preprint arXiv:2402.06659},
	year={2024}
}

@misc{gu2017badnets,
  title        = {BadNets: Identifying Vulnerabilities in the Machine Learning Model Supply Chain},
  author       = {Gu, Tianyu and Dolan-Gavitt, Brendan and Garg, Siddharth},
  year         = {2017},
  archivePrefix= {arXiv},
  eprint       = {1708.06733},
  primaryClass = {cs.CR}
}

@book{couillet2022random,
  title={Random matrix methods for machine learning},
  author={Couillet, Romain and Liao, Zhenyu},
  year={2022},
  publisher={Cambridge University Press}
}

@misc{EUAIActArt15,
  title        = {Article 15: Accuracy, Robustness and Cybersecurity (EU Artificial Intelligence Act)},
  howpublished = {\url{https://artificialintelligenceact.eu/article/15/}},
  note         = {Accessed 2025-10-01}
}

@misc{EUAIActRecital76,
  title        = {Recital 76 (EU Artificial Intelligence Act): Cybersecurity and data poisoning risks},
  howpublished = {\url{https://artificialintelligenceact.eu/recital/76/}},
  note         = {Accessed 2025-10-01}
}

@techreport{NIST_RMF,
  author       = {Elham Tabassi and others},
  title        = {Artificial Intelligence Risk Management Framework (AI RMF 1.0)},
  institution  = {National Institute of Standards and Technology},
  number       = {NIST AI 100-1},
  year         = {2023},
  url          = {https://nvlpubs.nist.gov/nistpubs/ai/nist.ai.100-1.pdf}
}

@techreport{NIST_GenAI_Profile,
  author       = {National Institute of Standards and Technology},
  title        = {AI Risk Management Framework: Generative AI Profile},
  institution  = {NIST},
  number       = {NIST AI 600-1},
  year         = {2024},
  url          = {https://nvlpubs.nist.gov/nistpubs/ai/NIST.AI.600-1.pdf}
}

@techreport{NIST_AML_Taxonomy,
  author       = {Vassilev, Apostol and others},
  title        = {Adversarial Machine Learning: A Taxonomy and Terminology of Attacks and Mitigations},
  institution  = {National Institute of Standards and Technology},
  number       = {NIST AI 100-2 (E2025)},
  year         = {2025},
  url          = {https://nvlpubs.nist.gov/nistpubs/ai/NIST.AI.100-2e2025.pdf}
}

@misc{MITRE_ATLAS,
  title        = {MITRE ATLAS: Adversarial Threat Landscape for Artificial-Intelligence Systems},
  howpublished = {\url{https://atlas.mitre.org/}},
  note         = {Accessed 2025-10-01}
}

@misc{OWASP_LLM_2025,
  title        = {OWASP GenAI/LLM Top 10 (2025): Data and Model Poisoning},
  howpublished = {\url{https://genai.owasp.org/llm-top-10/}},
  note         = {Accessed 2025-10-01}
}

@misc{OWASP_LLM_2023,
  title        = {OWASP Top 10 for LLM Applications (Archive 2023): Training Data Poisoning},
  howpublished = {\url{https://owasp.org/www-project-top-10-for-large-language-model-applications/Archive/0_1_vulns/Training_Data_Poisoning.html}},
  note         = {Accessed 2025-10-01}
}

@article{Gebru2021Datasheets,
  author       = {Gebru, Timnit and Morgenstern, Jamie and Vecchione, Briana and Vaughan, Jennifer Wortman and Wallach, Hanna and Daum{\'e} III, Hal and Crawford, Kate},
  title        = {Datasheets for Datasets},
  journal      = {Communications of the ACM},
  year         = {2021},
  volume       = {64},
  number       = {12},
  pages        = {86--92},
  doi          = {10.1145/3458723},
  url          = {https://dl.acm.org/doi/10.1145/3458723}
}

@inproceedings{Mitchell2019ModelCards,
  author       = {Mitchell, Margaret and Wu, Simone and Zaldivar, Andrew and Barnes, Parker and Vasserman, Lucy and Hutchinson, Ben and Spitzer, Elena and Raji, Inioluwa Deborah and Gebru, Timnit},
  title        = {Model Cards for Model Reporting},
  booktitle    = {Proceedings of the ACM Conference on Fairness, Accountability, and Transparency (FAccT)},
  year         = {2019},
  doi          = {10.1145/3287560.3287596},
  url          = {https://dl.acm.org/doi/10.1145/3287560.3287596}
}

@article{Arnold2019Factsheets,
  author       = {Arnold, Matthew and Bellamy, Rachel K. E. and others},
  title        = {FactSheets: Increasing Trust in AI Services through Supplier's Declarations of Conformity},
  journal      = {IBM Journal of Research and Development},
  year         = {2019},
  volume       = {63},
  number       = {4/5},
  doi          = {10.1147/JRD.2019.2942288},
  url          = {https://arxiv.org/abs/1808.07261}
}

@techreport{Brundage2018MaliciousUse,
  author       = {Brundage, Miles and Avin, Shahar and Clark, Jack and Toner, Helen and Eckersley, Peter and Garfinkel, Ben and Dafoe, Allan and Scharre, Paul and Zeitzoff, Thomas and Filar, Bobby and others},
  title        = {The Malicious Use of Artificial Intelligence: Forecasting, Prevention, and Mitigation},
  institution  = {Future of Humanity Institute et al.},
  year         = {2018},
  url          = {https://arxiv.org/abs/1802.07228}
}

@article{Schwartz2020GreenAI,
  author       = {Schwartz, Roy and Dodge, Jesse and Smith, Noah A. and Etzioni, Oren},
  title        = {Green AI},
  journal      = {Communications of the ACM},
  year         = {2020},
  volume       = {63},
  number       = {12},
  pages        = {54--63},
  doi          = {10.1145/3381831},
  url          = {https://cacm.acm.org/research/green-ai/}
}

@article{Hastie2019Surprises,
  title   = {Surprises in high-dimensional ridgeless least squares interpolation},
  author  = {Hastie, Trevor and Montanari, Andrea and Rosset, Saharon and Tibshirani, Ryan J.},
  journal = {arXiv preprint arXiv:1903.07571},
  year    = {2019}
}

@article{Dobriban2018Highdimridge,
  title={High-dimensional asymptotics of prediction: Ridge regression and classification},
  author={Dobriban, Edgar and Wager, Stefan},
  journal={Annals of Statistics},
  volume={46},
  number={1},
  pages={247--279},
  year={2018},
  publisher={Institute of Mathematical Statistics},
  doi={10.1214/17-AOS1554}
}

@article{Mei2019Generalization,
  title={Generalization error of random features and kernel methods: Universality, approximate message passing, and kernel ridge regression},
  author={Mei, Song and Montanari, Andrea and Nguyen, Phan-Minh},
  journal={Annals of Statistics},
  volume={47},
  number={6},
  pages={2797--2826},
  year={2019},
  publisher={Institute of Mathematical Statistics},
  doi={10.1214/19-AOS1780}
}

@inproceedings{Steinhardt2017Certified,
  title     = {Certified Defenses for Data Poisoning Attacks},
  author    = {Steinhardt, Jacob and Koh, Pang Wei and Liang, Percy},
  booktitle = {Advances in Neural Information Processing Systems},
  volume    = {30},
  year      = {2017}
}

@inproceedings{Jagielski2018RegressionPoisoning,
  title     = {Manipulating Machine Learning: Poisoning Attacks and Countermeasures for Regression Learning},
  author    = {Jagielski, Matthew and Oprea, Alina and Biggio, Battista and Liu, Chang and Nita-Rotaru, Cristina and Li, Bo},
  booktitle = {2018 IEEE Symposium on Security and Privacy (SP)},
  pages     = {19--35},
  year      = {2018},
  publisher = {IEEE}
}

@inproceedings{Tran2018SpectralSignatures,
  title     = {Spectral Signatures in Backdoor Attacks},
  author    = {Tran, Brandon and Li, Jerry and Madry, Aleksander},
  booktitle = {Advances in Neural Information Processing Systems},
  volume    = {31},
  pages     = {8000--8010},
  year      = {2018}
}

@inproceedings{Hayase2021Spectre,
  title     = {{SPECTRE}: Defending Against Backdoor Attacks Using Robust Statistics},
  author    = {Hayase, Jonathan and Kong, Weihao and Somani, Raghav and Oh, Sewoong},
  booktitle = {Proceedings of the 38th International Conference on Machine Learning},
  series    = {Proceedings of Machine Learning Research},
  volume    = {139},
  pages     = {4129--4139},
  year      = {2021},
  publisher = {PMLR}
}

@inproceedings{Xiang2023CBD,
  title     = {{CBD}: A Certified Backdoor Detector Based on Local Dominant Probability},
  author    = {Xiang, Zhen and Xiong, Zidi and Li, Bo},
  booktitle = {Advances in Neural Information Processing Systems},
  year      = {2023}
}

@article{Wang2020SmoothingBackdoor,
  title   = {On Certifying Robustness against Backdoor Attacks via Randomized Smoothing},
  author  = {Wang, Binghui and Cao, Xiaoyu and Jia, Jinyuan and Gong, Neil Zhenqiang},
  journal = {arXiv preprint arXiv:2002.11750},
  year    = {2020}
}

@inproceedings{Xie2021CRFL,
  title     = {{CRFL}: Certifiably Robust Federated Learning against Backdoor Attacks},
  author    = {Xie, Chulin and Chen, Minghao and Chen, Pin-Yu and Li, Bo},
  booktitle = {Proceedings of the 38th International Conference on Machine Learning},
  series    = {Proceedings of Machine Learning Research},
  volume    = {139},
  pages     = {11372--11382},
  year      = {2021},
  publisher = {PMLR}
}

@inproceedings{ManojBlum2021ExcessCapacity,
  title     = {Excess Capacity and Backdoor Poisoning},
  author    = {Manoj, Naren Sarayu and Blum, Avrim},
  booktitle = {Advances in Neural Information Processing Systems},
  volume    = {34},
  pages     = {20373--20384},
  year      = {2021}
}

@inproceedings{Zhang2024Orthogonality,
  title     = {Exploring the Orthogonality and Linearity of Backdoor Attacks},
  author    = {Zhang, Kaiyuan and Cheng, Siyuan and Shen, Guangyu and Tao, Guanhong and An, Shengwei and Makur, Anuran and Ma, Shiqing and Zhang, Xiangyu},
  booktitle = {2024 IEEE Symposium on Security and Privacy (SP)},
  year      = {2024},
  doi       = {10.1109/SP54263.2024.00225},
  publisher = {IEEE}
}

@inproceedings{Xian2023Adaptability,
  title     = {Understanding Backdoor Attacks through the Adaptability Hypothesis},
  author    = {Xian, Xun and Bi, Xuan and Hong, Mingyi and Wang, Ganghua and Ding, Jie},
  booktitle = {Proceedings of the 40th International Conference on Machine Learning},
  series    = {Proceedings of Machine Learning Research},
  volume    = {202},
  year      = {2023},
  publisher = {PMLR}
}

@inproceedings{Goldwasser2022UndetectableBackdoors,
  title     = {Planting Undetectable Backdoors in Machine Learning Models},
  author    = {Goldwasser, Shafi and Kim, Michael P. and Vaikuntanathan, Vinod and Zamir, Or},
  booktitle = {2022 IEEE 63rd Annual Symposium on Foundations of Computer Science (FOCS)},
  pages     = {931--942},
  year      = {2022},
  publisher = {IEEE}
}

@misc{szegedy_intriguing_2014,
    title = {Intriguing properties of neural networks},
    url = {http://arxiv.org/abs/1312.6199},
    doi = {10.48550/arXiv.1312.6199},
    urldate = {2025-11-26},
    publisher = {arXiv},
    author = {Szegedy, Christian and Zaremba, Wojciech and Sutskever, Ilya and Bruna, Joan and Erhan, Dumitru and Goodfellow, Ian and Fergus, Rob},
    month = feb,
    year = {2014},
    note = {arXiv:1312.6199 [cs]},
}

@misc{ghosh_scalable_2021,
    title = {Scalable logistic regression with crossed random effects},
    url = {http://arxiv.org/abs/2105.13747},
    doi = {10.48550/arXiv.2105.13747},
    urldate = {2025-11-26},
    publisher = {arXiv},
    author = {Ghosh, Swarnadip and Hastie, Trevor and Owen, Art B.},
    month = dec,
    year = {2021},
    note = {arXiv:2105.13747 [stat]},
}

@misc{bellio_consistent_2025,
    title = {Consistent and {Scalable} {Composite} {Likelihood} {Estimation} of {Probit} {Models} with {Crossed} {Random} {Effects}},
    url = {http://arxiv.org/abs/2308.15681},
    doi = {10.48550/arXiv.2308.15681},
    urldate = {2025-11-26},
    publisher = {arXiv},
    author = {Bellio, Ruggero and Ghosh, Swarnadip and Owen, Art B. and Varin, Cristiano},
    month = apr,
    year = {2025},
    note = {arXiv:2308.15681 [stat]},
}

@misc{zhou_learning_2025,
    title = {Learning to {Poison} {Large} {Language} {Models} for {Downstream} {Manipulation}},
    url = {http://arxiv.org/abs/2402.13459},
    doi = {10.48550/arXiv.2402.13459},
    urldate = {2026-01-05},
    publisher = {arXiv},
    author = {Zhou, Xiangyu and Qiang, Yao and Zade, Saleh Zare and Roshani, Mohammad Amin and Khanduri, Prashant and Zytko, Douglas and Zhu, Dongxiao},
    month = may,
    year = {2025},
    note = {arXiv:2402.13459 [cs]
version: 3},
}

@misc{li_badedit_2024,
    title = {{BadEdit}: {Backdooring} large language models by model editing},
    shorttitle = {{BadEdit}},
    url = {http://arxiv.org/abs/2403.13355},
    doi = {10.48550/arXiv.2403.13355},
    urldate = {2026-01-04},
    publisher = {arXiv},
    author = {Li, Yanzhou and Li, Tianlin and Chen, Kangjie and Zhang, Jian and Liu, Shangqing and Wang, Wenhan and Zhang, Tianwei and Liu, Yang},
    month = mar,
    year = {2024},
    note = {arXiv:2403.13355 [cs]},
}

@misc{yao_poisonprompt_2023,
    title = {{PoisonPrompt}: {Backdoor} {Attack} on {Prompt}-based {Large} {Language} {Models}},
    shorttitle = {{PoisonPrompt}},
    url = {http://arxiv.org/abs/2310.12439},
    doi = {10.48550/arXiv.2310.12439},
    urldate = {2026-01-04},
    publisher = {arXiv},
    author = {Yao, Hongwei and Lou, Jian and Qin, Zhan},
    month = dec,
    year = {2023},
    note = {arXiv:2310.12439 [cs]},
}

@misc{zhang_instruction_2024,
    title = {Instruction {Backdoor} {Attacks} {Against} {Customized} {LLMs}},
    url = {http://arxiv.org/abs/2402.09179},
    doi = {10.48550/arXiv.2402.09179},
    urldate = {2026-01-04},
    publisher = {arXiv},
    author = {Zhang, Rui and Li, Hongwei and Wen, Rui and Jiang, Wenbo and Zhang, Yuan and Backes, Michael and Shen, Yun and Zhang, Yang},
    month = may,
    year = {2024},
    note = {arXiv:2402.09179 [cs]},
}

@misc{zhou_aspire_2024,
    title = {{ASPIRe}: {An} {Informative} {Trajectory} {Planner} with {Mutual} {Information} {Approximation} for {Target} {Search} and {Tracking}},
    shorttitle = {{ASPIRe}},
    url = {http://arxiv.org/abs/2403.01674},
    doi = {10.48550/arXiv.2403.01674},
    urldate = {2026-01-04},
    publisher = {arXiv},
    author = {Zhou, Kangjie and Wu, Pengying and Su, Yao and Gao, Han and Ma, Ji and Liu, Hangxin and Liu, Chang},
    month = mar,
    year = {2024},
    note = {arXiv:2403.01674 [cs]},
}

@misc{he_multi-faceted_2025,
    title = {Multi-{Faceted} {Studies} on {Data} {Poisoning} can {Advance} {LLM} {Development}},
    url = {http://arxiv.org/abs/2502.14182},
    doi = {10.48550/arXiv.2502.14182},
    urldate = {2026-01-04},
    publisher = {arXiv},
    author = {He, Pengfei and Xing, Yue and Xu, Han and Xiang, Zhen and Tang, Jiliang},
    month = feb,
    year = {2025},
    note = {arXiv:2502.14182 [cs]
version: 1},
}

@misc{ge_when_2024,
    title = {When {Backdoors} {Speak}: {Understanding} {LLM} {Backdoor} {Attacks} {Through} {Model}-{Generated} {Explanations}},
    shorttitle = {When {Backdoors} {Speak}},
    url = {http://arxiv.org/abs/2411.12701},
    doi = {10.48550/arXiv.2411.12701},
    urldate = {2026-01-04},
    publisher = {arXiv},
    author = {Ge, Huaizhi and Li, Yiming and Wang, Qifan and Zhang, Yongfeng and Tang, Ruixiang},
    month = nov,
    year = {2024},
    note = {arXiv:2411.12701 [cs]
version: 1},
}

@article{deng_mnist_2012,
	title = {The {MNIST} {Database} of {Handwritten} {Digit} {Images} for {Machine} {Learning} {Research} [{Best} of the {Web}]},
	volume = {29},
	issn = {1558-0792},
	url = {https://ieeexplore.ieee.org/document/6296535},
	doi = {10.1109/MSP.2012.2211477},
	number = {6},
	urldate = {2025-10-01},
	journal = {IEEE Signal Processing Magazine},
	author = {Deng, Li},
	month = nov,
	year = {2012},
	pages = {141--142},
}

@book{wainwright_high-dimensional_2019,
	address = {Cambridge},
	series = {Cambridge {Series} in {Statistical} and {Probabilistic} {Mathematics}},
	title = {High-{Dimensional} {Statistics}: {A} {Non}-{Asymptotic} {Viewpoint}},
	isbn = {978-1-108-49802-9},
	shorttitle = {High-{Dimensional} {Statistics}},
	url = {https://www.cambridge.org/core/books/highdimensional-statistics/8A91ECEEC38F46DAB53E9FF8757C7A4E},
	urldate = {2025-09-29},
	publisher = {Cambridge University Press},
	author = {Wainwright, Martin J.},
	year = {2019},
	doi = {10.1017/9781108627771},
}

@misc{carlini_poisoning_2022,
	title = {Poisoning and {Backdooring} {Contrastive} {Learning}},
	url = {http://arxiv.org/abs/2106.09667},
	doi = {10.48550/arXiv.2106.09667},
	urldate = {2025-09-24},
	publisher = {arXiv},
	author = {Carlini, Nicholas and Terzis, Andreas},
	month = mar,
	year = {2022},
	note = {arXiv:2106.09667 [cs]},
}

@inproceedings{shafahi_poison_2018,
	title = {Poison {Frogs}! {Targeted} {Clean}-{Label} {Poisoning} {Attacks} on {Neural} {Networks}},
	volume = {31},
	url = {https://papers.nips.cc/paper_files/paper/2018/hash/22722a343513ed45f14905eb07621686-Abstract.html},
	urldate = {2025-09-24},
	booktitle = {Advances in {Neural} {Information} {Processing} {Systems}},
	publisher = {Curran Associates, Inc.},
	author = {Shafahi, Ali and Huang, W. Ronny and Najibi, Mahyar and Suciu, Octavian and Studer, Christoph and Dumitras, Tudor and Goldstein, Tom},
	year = {2018},
}

@misc{turner_label-consistent_2019,
	title = {Label-{Consistent} {Backdoor} {Attacks}},
	url = {http://arxiv.org/abs/1912.02771},
	doi = {10.48550/arXiv.1912.02771},
	urldate = {2025-09-24},
	publisher = {arXiv},
	author = {Turner, Alexander and Tsipras, Dimitris and Madry, Aleksander},
	month = dec,
	year = {2019},
	note = {arXiv:1912.02771 [stat]},
}

@book{vershynin_high-dimensional_2018,
	address = {Cambridge},
	series = {Cambridge series in statistical and probabilistic mathematics},
	title = {High-dimensional probability: an introduction with applications in data science},
	isbn = {978-1-108-41519-4},
	shorttitle = {High-dimensional probability},
	language = {eng},
	number = {47},
	publisher = {University Press},
	author = {Vershynin, Roman},
	year = {2018},
}

@book{bai_spectral_2010,
	address = {New York, NY},
	series = {Springer {Series} in {Statistics}},
	title = {Spectral {Analysis} of {Large} {Dimensional} {Random} {Matrices}},
	copyright = {https://www.springernature.com/gp/researchers/text-and-data-mining},
	isbn = {978-1-4419-0660-1 978-1-4419-0661-8},
	url = {https://link.springer.com/10.1007/978-1-4419-0661-8},
	language = {en},
	urldate = {2025-07-23},
	publisher = {Springer New York},
	author = {Bai, Zhidong and Silverstein, Jack W.},
	year = {2010},
	doi = {10.1007/978-1-4419-0661-8},
	note = {ISSN: 0172-7397},
}

@book{couillet_random_2022,
	edition = {1},
	title = {Random {Matrix} {Methods} for {Machine} {Learning}},
	copyright = {https://www.cambridge.org/core/terms},
	isbn = {978-1-009-12849-0 978-1-009-12323-5},
	url = {https://www.cambridge.org/core/product/identifier/9781009128490/type/book},
	language = {en},
	urldate = {2025-05-28},
	publisher = {Cambridge University Press},
	author = {Couillet, Romain and Liao, Zhenyu},
	month = jul,
	year = {2022},
	doi = {10.1017/9781009128490},
}

\newpage
\onecolumn
\clearpage
\appendix
\thispagestyle{empty}

% Supplementary material: To improve readability, you must use a single-column format for the supplementary material.
\onecolumn

\appendix
\clearpage
\section{Appendix}

\subsection{Proof of Results}

In this section we provide full proofs of the results for Theorem 3.1 and Proposition 4.4.
The proofs are done using random matrix theory, and particularly the resolvent techniques put forward by \cite{couillet2022random}. 

We use the following notation from \cite{couillet2022random}:

\begin{definition}
	For $\X, \Y \in \RR^{n \times n}$ two random or deterministic matrices, we write
	\[
	\X \longleftrightarrow \Y,
	\]
	if, for all $\A \in \RR^{n \times n}$ and $\ba, \bb \in \RR^n$ of unit norms (respectively, operator and Euclidean), we have the simultaneous results
	\[
	\frac{1}{n} \, \mathrm{tr} \, \A(\X - \Y) \to 0, \quad \ba^\top (\X - \Y) \bb \to 0, \quad \|\EE[\X - \Y]\| \to 0,
	\]
	where, for random quantities, the convergence is either in probability or almost surely.
\end{definition}

We then prove the following two deterministic equivalent lemmas, allowing us to understand the spiked resolvent

\begin{lemma}
	\label{q1_statistic_lemma}
	Suppose $\X$ is a $\RR^{p \times n}$ matrix, such that each entry $X_{i, j} \sim N(0,1)$ 
	is an independent and identically distributed Gaussian random variable.
	Now $\Z = \X + \tau \sqrt{n} \ba\bb^\top$, where $\ba$ and $\bb$ are of unit norm in $\RR^p$ and $\RR^n$ respectively.

	The spiked resolvent $\Q_1(z) = (\frac1n \Z \Z^\top - z \I_p)^{-1}$ satisfies the deterministic equivalent
	$$\Q_1 \longleftrightarrow  \bar{\Q}_1 := m(z)\I_p -  m(z)\left( 1 - \frac{1}{1 + \tau^2 (1 + zm(z))}\right)\ba \ba^\top $$
\end{lemma}

\begin{lemma}
	\label{lemma:q1_squared_det_equiv}
	Under the same assumptions as above, the square of the spiked resolvent $\Q_1^2$ satisfies the deterministic equivalent
	$$\Q_1^2 \longleftrightarrow  m'(z)\I_p + \left(\frac{m'(z) (\tau^2 + 1) - m(z)^2 \tau^2}{\big(1 + \tau^2(1 + zm(z)) \big)^2} - m'(z)\right)\ba \ba^\top$$
	In particular, for
	$$\tilde{\Q}_1 := \left(\frac1n \Z^\top \Z - z \I_n\right)^{-1}$$
	Then,
	$$\tilde{\Q}_1^2 \longleftrightarrow \tilde{m}'(z) \I_n + \left(\frac{(c^{-1} \tau^2 + 1)\tilde{m}'(z) - \tilde{m}(z)^2 \tau^2 c^{-1}}{\big(1 + c^{-1} \tau^2 (1 + z \tilde{m}(z)) \big)^2} - \tilde{m}'(z)\right)\bb \bb^\top$$
\end{lemma}

Where $\tilde{m}(z)$ is the Stieltjes transform corresponding to the Gram matrix resolvent $\tilde{\Q}(z)$, and satisfies the relations
$\tilde{m}(z) = cm(z) - \frac{1-c}{z}$, and $\tilde{m}(z) = c^{-1}m_{\text{flip}}(c^{-1}z)$,
where $m_{\text{flip}}(z)$ is the Stieltjes transform obtained by simply substituting $c^{-1}$ for $c$ in the
definition of $m(z)$.

\subsection{Proof of Proposition 4.4}
\begin{proof}
We consider first the expectation, we claim that $\EE[\hat \bbeta^\top \ba] \rightarrow C \bv^\top \ba$

Then we write,
\[
\EE[\ba^\top \hat \bbeta] = \EE\left[ \frac{1}{n} \ba^\top \Q_1 \widetilde{\Z}\widetilde \w \right] = \sum_{i=1}^n \frac{1}{n} \EE \left[ \ba^\top \Q_1 \widetilde \z_i \widetilde w_i \right]
\]

Then to handle the dependence between $\Q_1$ and $\widetilde \z_i$ we employ a trick commonly used in the textbook \citep{couillet2022random}
of removing the $\widetilde \z_i$ term from $\Q_1$. This amounts to a rank-1 perturbation of $\widetilde \Z \widetilde \Z^\top$, which 
we can then understand the behaviour on $\Q_1$ through the Sherman–Morrison formula.

Since $\Q_1 = \left(\frac{1}{n} \widetilde \Z \widetilde \Z^\top + \lambda \I\right)^{-1} = \left(\frac{1}{n} \sum_{i=1}^n \widetilde \z_i \widetilde \z_i^\top + \lambda \I\right)^{-1}$, we define the ``leave-one-out''
resolvent

$$\Q_1^{-i} = \left(\frac{1}{n} \sum_{j\neq i} \widetilde \z_j \widetilde \z_j^\top + \lambda \I\right)^{-1}$$

And using the Sherman–Morrison formula, we compute:

$$\Q_1 \widetilde \z_i = \frac{\Q_1^{-i} \widetilde \z_i}{1 + \frac{1}{n} \widetilde \z_i^\top \Q_1^{-i}\widetilde \z_i}$$

Then we claim that the denominator here satisfies\footnote{In fact we can take higher powers here, but the variance is sufficient for our purposes.}
\begin{equation}
\label{eqn:Q_trace_denom}
    \EE \left| \frac{1}{1 + \frac{1}{n} \widetilde \z_i^\top \Q_1^{-i}\widetilde \z_i} - \frac{1}{1 + cm(-\lambda)} \right|^2 = \frac{1}{n}
\end{equation}

To see this we use two facts
\begin{enumerate}
    \item For $\x_i \sim \mathcal{N}(0, \I_p)$ and $\A$ independent of $\x_i$ with $\| \A \| = O(1)$, then 
    \begin{equation}
    \label{eqn:quad_form_trace}
        \EE | \tfrac{1}{n}\x_i^\top \A \x_i - \tr \A|^2 = O(n^{-1}) 
    \end{equation}
    \item \begin{equation}
    \label{eqn:trace_q_lambda}
        \tfrac1p \tr \Q_1(-\lambda) = m(-\lambda) + O(n^{-1})
    \end{equation}
\end{enumerate}
Equation \ref{eqn:quad_form_trace} can be seen by an explicit calculation of the variance, or as a special case of \cite{couillet2022random} Lemma 2.9. And Equation \ref{eqn:trace_q_lambda} can be seen by the Sherman Morrison formula, as $\Q_1$ is a rank 3 perturbation of the Wishart matrix resolvent. Denoting this by $\tfrac{1}{n}\tr\tilde{\Q}_\lambda = (\tfrac{1}{n}\X \X^\top + \lambda \I_p)^{-1}$, it is a standard fact that $\tilde{\Q}_\lambda = m(-\lambda) + O(n^{-1})$, and then by the Woodbury Lemma we see that finite rank perturbations don't affect the normalised trace.

Following from these facts then, we have that 
\begin{align*}
\EE \left| \frac{1}{1 + \frac{1}{n} \widetilde \z_i^\top \Q_1^{-i}\widetilde \z_i} - \frac{1}{1 + cm(-\lambda)} \right|^2 
&\leq C\left(\EE \left| \frac{1}{n} \widetilde \z_i^\top \Q_1^{-i}\widetilde \z_i - \frac1n \x_i^\top \Q_1^{-i} \x
\right|^2 + \EE \left|\frac1n \x_i^\top \Q_1^{-i}\x - cm(-\lambda)
\right|^2\right) \\
&= O(n^{-1})
\end{align*}
    
Since as $\widetilde \Z = \X + \bv \rr^\top$, then $\widetilde \z_i = \x_i + \bv r_i$, and in particular $\|\widetilde \z_i - \x_i\| = O(1)$

Returning now to our calculation, we have
\[
\EE[ \ba^\top \hat \bbeta ] = \sum_{i=1}^n \frac1n \EE\frac{\ba^\top\Q_1^{-i} \widetilde \z_i \widetilde w_i}{1 + \frac{1}{n} \widetilde \z_i^\top \Q_1^{-i}\widetilde \z_i}
= \sum_{i=1}^n \frac1n \EE\frac{\ba^\top\Q_1^{-i} \widetilde \z_i \widetilde w_i}{1 + cm(-\lambda)} + O(n^{-1/2})
\]
This comes from using equation \ref{eqn:Q_trace_denom} along a H\"older inequality and the fact that since $\widetilde \z_i$ is independent of $\Q_1^{-i}$, the variance of the numerator is bounded, specifically we have $\EE \left[ (\ba^\top \Q_1^{-i} \widetilde \z_i \widetilde w_i)^2 \ | \ \Q_1^{-i}\right] \leq \ba^\top (\Q_1^{-i})^2 \ba\leq \| \ba \|^2/\lambda^2$ (since $\widetilde w_i \leq 1$).

Following from these we immediately see on breaking $\widetilde \z_i = \x_i + \bv r_i$ that the $\x_i$ term vanishes, as it is mean $0$ and independent. Hence,

\begin{align*}
\EE[ \ba^\top \hat \bbeta ] &= \frac{1}{1 + cm(-\lambda)}\frac1n\sum_{i=1}^n \EE \ba^\top \Q_1^{-i} \bv r_i \widetilde w_i + O(n^{-1/2}) \\
&= \frac{1}{1 + cm(-\lambda)}\frac1n \ba^\top \bar \Q_1 \bv \rr^\top \widetilde \w + O(n^{-1/2})
\end{align*}

Here we have used Lemma \ref{q1_statistic_lemma} for the deterministic equivalent of the resolvent. 

Finally, we evaluate this limit. From the setup of our problem we see that, $\frac{1}{n} \rr^\top \widetilde \w \stackrel{a.s.}{\rightarrow} \frac{1}{2} \theta (1-\theta)$ and $\frac{1}{n} \| \rr\|^2 \stackrel{a.s.}{\rightarrow} \frac{\theta}{2}(1- \frac{\theta}{2})$.

Hence, expanding out the deterministic equivalent we get
\[
\EE[ \ba^\top \hat \bbeta ] = \frac{m(-\lambda) \theta (1 - \theta) \ba^\top \bv }{(1 + cm(-\lambda))(2+\| \bv \|^2 \theta(1 - \tfrac{\theta}{2})(1 - \lambda m(-\lambda)))}.
\]

To show convergence, we then use a concentration of measure argument to say that $\ba^\top \hat \bbeta \rightarrow \EE [ \ba^\top \hat \bbeta]$. For this we use the Burkholder inequality for martingale differences.
\begin{lemma}[Burkholder inequality \cite{bai_spectral_2010}, Lemma 2.13]
Let $\{X_i\}_{i=1}^{\infty}$ be a martingale difference sequence for the filtration $\{\mathcal F_i\}$ and denote by $\mathbb E_k$ the conditional expectation with respect to $\mathcal F_k$.
Then, for $k\ge2$, there exists a constant $C_k$ depending only on $k$ such that

\[
\mathbb E\Bigl|\sum_{i=1}^{n} X_i\Bigr|^{k}
\;\le\; C_k\left(
\mathbb E\Bigl[\sum_{i=1}^{n}\mathbb E_{i-1}\!\left(|X_i|^{2}\right)\Bigr]^{k/2}
+\sum_{i=1}^{n}\mathbb E\!\left(|X_i|^{k}\right)
\right).
\]

\end{lemma}

The utility of this comes by letting the filtration $\mathcal{F}_i = \sigma(\widetilde \z_1, \ldots \widetilde \z_i)$, and then writing \[
\ba^\top \hat \bbeta - \EE [ \ba^\top \hat \bbeta ] = \sum_i (\EE_i - \EE_{i-1})(\ba^\top \hat \bbeta) = \sum_i (\EE_i - \EE_{i-1})(\ba^\top \hat \bbeta - \ba^\top \hat \bbeta_{-i})
\]
Where here we let $\hat \bbeta_{-i}$ be the estimator constructed from missing out data point $(\widetilde \z_i, \widetilde \w_i)$. This gives us a concrete way to use the intuition that if we leave out one datapoint it shouldn't change the result, since by the Burkholder inequality, if we can control the moments of $\ba^\top \hat \bbeta - \ba^\top \hat \bbeta_{-i}$ then can control the fluctuations from the mean. In particular, if we choose $k=4$ in Burkholder, then we can show that the 4th moment of the left side is $O(n^{-2})$, and hence we have absolute convergence.

Using again Sherman-Morrison, we then get:
\[
\ba^\top \hat \bbeta - \ba^\top \hat \bbeta_{-i} = \ba^\top \left(\Q_1^{-i} - \frac1n \frac{\Q_1^{-i} \widetilde \z_i \widetilde \z_i^\top \Q_1^{-i}}{1 + \frac{1}{n}\widetilde\z_i\Q_1^{-i} \widetilde \z_i} \right)\left( \frac1n \sum_{j \neq i} \widetilde \z_j \widetilde w_j + \frac1n  \widetilde \z_i \widetilde w_i \right) - \ba^\top \hat \bbeta_{-i}
\]

The first term from each bracket then multiplies to cancel exactly the $ \ba^\top \hat \bbeta_{-i}$ term. We then need to argue that 4th moments of the remaining terms are small to use Burkholder.
\begin{enumerate}
    \item $\frac1n \ba^\top \Q_1^{-i}\widetilde \z_i \widetilde w_i$. To bound this term, we use independence of $\widetilde \z_i$ and $\Q_1^{-i}$ to see that it is conditionally normal, with mean and variance depending on $\| \Q_1^{-i}\ba\|$. By boundedness of $\Q_1^{-i}$ this term then has for example $$\EE\left[ \left| \frac1n \ba^\top \Q_1^{i}\widetilde \z_i \widetilde w_i\right|^4 \right] = O(n^{-4})$$
    \item To bound the term \[
    \EE \left[ \left| \frac{1}{n^2} \ba^\top \frac{\Q_1^{-i} \widetilde \z_i \widetilde \z_i^\top \Q_1^{-i}}{1 + \frac{1}{n}\widetilde\z_i\Q_1^{-i} \widetilde \z_i} \widetilde \z_i \widetilde w_i \right|^4 \right],
    \] we use H\"older's inequality, and positive definiteness of the denominator to get \[
    \leq \frac{1}{n^4} \EE\left[\left| \ba^\top \Q_1^{-i} \widetilde \z_i \right|^8 \right]^{1/2}
    \EE\left[\left| \frac1n \widetilde \z_i ^\top \Q_1^{-i} \widetilde \z_i \right|^8 \right]^{1/2}.
    \]
    The first term is $O(1)$ by the same conditional normal argument as above, while the second term is $O(1)$ by a simple norm bound, and hence we again get the required $O(n^{-4})$.
    \item The final term is 
    \begin{align*}
    &\EE \left[ \left| \frac{1}{n^2} \ba^\top \frac{\Q_1^{-i} \widetilde \z_i \widetilde \z_i^\top \Q_1^{-i}}{1 + \frac{1}{n}\widetilde\z_i\Q_1^{-i} \widetilde \z_i}\left(\frac1n \sum_{j \neq i} \widetilde \z_j \widetilde w_j \right)\right|^4 \right] \\
    \leq&  \frac{1}{n^4} \EE\left[\left| \ba^\top \Q_1^{-i} \widetilde \z_i \right|^8 \right]^{1/2} \EE\left[ \left| \widetilde \z_i^\top \hat{\bbeta}_{-i} \right|^8 \right]^{1/2}.
    \end{align*}
    Where we have again used H\"older's inequality and that $\hat \bbeta_{-i} = \frac1n \Q_1^{-i} \sum_{j \neq i} \widetilde \z_j \widetilde w_j$. We can then use a conditional normal argument again, since $\widetilde \z_i$ is independent from $\hat \bbeta_{-i}$, and note simply that $\| \hat \bbeta_{-i} \|^2$ is absolutely bounded since we may rearrange
    \[
    \|\hat\bbeta\|^2=\frac{1}{n}\,\widetilde\w^\top\!\left(z\,\tilde\Q_1(z)^2+\tilde\Q_1(z)\right)\widetilde\w
    \]
    for \(\tilde\Q_1(z)=(\frac1n\widetilde\X^\top\widetilde\X-z\I_n)^{-1}\).
\end{enumerate}
Hence, by Burkholder $\EE[ | \ba^\top \hat \bbeta  - \EE\ba^\top \hat \bbeta |^4 ] = O(n^{-2})$, and in particular \[
\ba^\top \hat \bbeta \stackrel{a.s.}{\rightarrow} \EE[\ba^\top \hat \bbeta] = \frac{m(-\lambda) \theta (1 - \theta) \ba^\top \bv }{(1 + cm(-\lambda))(2+\| \bv \|^2 \theta(1 - \tfrac{\theta}{2})(1 - \lambda m(-\lambda)))}
\]
\end{proof}

\subsection{Proof of Theorem 3.1}
\begin{proof}
    
Recall the setup,
\[
\widetilde\X=\X+\bv \rr^\top,\qquad
\widetilde\w=\y+2\rr,\qquad
\rr=\bu-\tfrac{\theta}{2}\one ,
\]
so that
\[
\frac{1}{n}\|\rr\|^2 \asto s:=\frac{\theta}{2}\Big(1-\frac{\theta}{2}\Big),\qquad
\frac{1}{n}\y^\top\rr \asto -\frac{\theta}{2}.
\]

Then by Proposition 4.4 we have that \(\hat\bbeta^\top\bv\) concentrates to a constant, which we identify as $\mu$. The randomness in
\(\hat\bbeta^\top(\x_0+\bv)\) then comes from \(\x_0\). Conditionally on \(\hat\bbeta\),
\(\hat\bbeta^\top \x_0 \sim \NN(0,\|\hat\bbeta\|^2)\), hence the asymptotic variance is
\(\sigma^2=\lim \|\hat\bbeta\|^2\).

Let \(z=-\lambda\), \(\Q_1(z)=(\frac1n\widetilde\X\widetilde\X^\top-z\I_p)^{-1}\),
\(\tilde\Q_1(z)=(\frac1n\widetilde\X^\top\widetilde\X-z\I_n)^{-1}\). Using
\(\frac{1}{n}\widetilde\X^\top \Q_1(z)^2 \widetilde\X= z\,\tilde\Q_1(z)^2+\tilde\Q_1(z)\),
\[
\|\hat\bbeta\|^2=\frac{1}{n}\,\widetilde\w^\top\!\left(z\,\tilde\Q_1(z)^2+\tilde\Q_1(z)\right)\widetilde\w\Big|_{z=-\lambda}.
\]

Set the sample‑space spike direction \(\bar\bb = \rr/\|\rr\|\),
\[
\tau^2:=\|\bv\|^2 s,\qquad a:=c^{-1}\tau^2,
\qquad B(z):=1+a\big(1+z\tilde m(z)\big),\qquad S(z):=\tilde m(z)+z\tilde m'(z).
\]
By Lemmas~\ref{q1_statistic_lemma}–\ref{lemma:q1_squared_det_equiv},
\[
\tilde\Q_1(z)\ \Longleftrightarrow\ \tilde m(z)\!\left(\I_n-\Big(1-\frac{1}{B(z)}\Big)\bar\bb\bar\bb^\top\right),
\]
\[
\tilde\Q_1(z)^2\ \Longleftrightarrow\ \tilde m'(z)\I_n+
\Big(T(z)-\tilde m'(z)\Big)\bar\bb\bar\bb^\top,\quad
T(z):=\frac{(a+1)\tilde m'(z)-a\,\tilde m(z)^2}{B(z)^2}.
\]
A direct algebraic check yields
\[
z\big(T(z)-\tilde m'(z)\big)-\tilde m(z)\!\left(1-\frac{1}{B(z)}\right)
= S(z)\!\left(\frac{a+1}{B(z)^2}-1\right).
\]

Two scalar Law of Large Numbers (LLN) give
\[
\frac{1}{n}\|\widetilde\w\|^2 \asto 1-\theta^2,\qquad
\frac{1}{n}\big(\widetilde\w^\top\bar\bb\big)^2
=\frac{1}{n}\frac{(\widetilde\w^\top \rr)^2}{\|\rr\|^2}
\asto \frac{\big(\EE[r_i\widetilde w_i]\big)^2}{\EE[r_i^2]}
= \frac{\theta(1-\theta)^2}{\,2-\theta\,}.
\]
Combining the deterministic equivalents with the LLNs,
\[
\frac{1}{n}\,\widetilde\w^\top\!\left(z\,\tilde\Q_1^2+\tilde\Q_1\right)\widetilde\w
\ \Longrightarrow\
S(z)\!\left[
\frac{1}{n}\|\widetilde\w\|^2
+\left(\frac{a+1}{B(z)^2}-1\right)\frac{1}{n}\big(\widetilde\w^\top\bar\bb\big)^2
\right].
\]
Evaluating at \(z=-\lambda\) gives the variance
\begin{equation}
\sigma^2=\Big(\tilde m(-\lambda)-\lambda \tilde m'(-\lambda)\Big)
\left[
(1-\theta^2)
+\left(\frac{1+a}{\big(1+a\,(1-\lambda \tilde m(-\lambda))\big)^2}-1\right)
\frac{\theta(1-\theta)^2}{\,2-\theta\,}
\right],
\end{equation}
with \(a=c^{-1}\tau^2\) and \(\tau^2=\|\bv\|^2\,\frac{\theta}{2}\!\left(1-\frac{\theta}{2}\right)\).

\begin{remark}[Unregularized limit]
If \(\lambda\to0\) with \(c<1\), then \(S(-\lambda)\to\frac{c}{1-c}\) and
\(B(-\lambda)\to 1+a c=1+\tau^2\). Hence
\[
\sigma_0^2=\frac{c}{1-c}\left[
(1-\theta^2)
+\left(\frac{1+c^{-1}\tau^2}{(1+\tau^2)^2}-1\right)
\frac{\theta(1-\theta)^2}{\,2-\theta\,}
\right].
\]
\end{remark}

\end{proof}

\subsection{Proof of Lemma \ref{q1_statistic_lemma}}

\begin{proof}
	The main idea behind the proof is we will establish that $\EE \Q_1 = \bar{\Q}_1 + o_{\|\cdot\|}(1)$, where $o_{\| \cdot \|}(1)$ is a matrix with operator norm converging to $0$ as $n, p \rightarrow \infty$. We then use a concentration argument to see that $\Q_1 = \bar\Q_1 + o_{\| \cdot \|}(1)$, where the difference here converges to $0$ in operator norm in probability.

	To do this, we will first write $\Q_1$ in a more convenient form.

	We have that $\Z = \X + \tau \sqrt{n} \ba \bb^\top$, and hence 
	
	$$\frac1n \Z \Z^\top = \frac1n ( \X + \tau \sqrt{ n}  \ba \bb^\top) ( \X + \tau  \sqrt{n}  \ba \bb^\top)^\top$$ 
	$$= \frac1n \X \X^\top + \frac{\tau}{\sqrt{n}}  \ba \bb^\top \X^\top + \frac{\tau}{\sqrt{n}}  \X \bb \ba^\top + \tau^2 \ba \ba ^\top$$

	Then we seek to isolate the low rank component of this by defining block matrices $\U, \V \in \RR^{p \times 3}$ such that $\frac 1n \Z \Z^\top = \frac 1n \X \X^\top + \U \V^\top$. We have some freedom in this in how we normalise
	the entries of $\U$ and $\V$, so we will choose a normalisation so that each entry of $\U$ and $\V$ has size $O(1)$

	So making the choice:
	$$\U = 
	\begin{bmatrix}
	\tau \ba & \frac{1}{\sqrt n}\X \bb & \tau \ba	
	\end{bmatrix}
	$$

	$$\V = 
	\begin{bmatrix}
	\frac{1}{\sqrt n} \X \bb& \tau \ba & \tau \ba	
	\end{bmatrix}
$$

We now have that $\frac1n \Z \Z^\top = \frac1n \X \X^\top + \U \V^\top$

Hence we may use the Woodbury formula to write

$$\Q_1(z) = \Q_0(z) - \Q_0(z) \U \left( \I_3 + \V^\top \Q_0(z) \U\right)^{-1} \V^\top \Q_0(z)$$

Where $\I_3$ is the $3 \times 3$ identity matrix.

Now defining, $\A = \left( \I_3 + \V^\top \Q_0(z) \U\right)^{-1}$, then we will argue that we can replace $\A$ with a deterministic matrix $\bar{\A}$.
Since $\A \in \RR^{3 \times 3}$, we may take limits in each entry of $\V^\top \Q_0 \U$ before inverting to get $\bar{\A}$, and from the finite dimensionality we then have that
the operator norm $\| \A - \bar{\A}\| \rightarrow 0$.

Our problem term in the expansion of $\Q_1$ can then be simplified,

$$\EE \left[ \Q_0 \U \A \V^\top \Q_0 \right] = \EE \left[ \Q_0 \U \bar{\A} \V^\top \Q_0 \right] + \EE \left[ \Q_0 \U (\A - \bar{\A}) \V^\top \Q_0 \right]$$ 
$$= \EE \left[ \Q_0 \U \bar{\A} \V^\top \Q_0 \right] + o_{\|\cdot\|}(1) $$

Since $\|\Q_0\|$, $\|\U\|$, and $\|\V\|$ are all $O(1)$ almost surely. (The operator norm of $\frac{1}{\sqrt{n}}\X$ has finite $\limsup$).

To actually compute the limit of $\V^\top \Q_0 \U$ we use the following almost sure limit identities where $\ba$ and $\bb$ represent unit vectors of the appropriate dimension

\begin{enumerate}
	\item $$\ba^\top \Q_0(z) \bb \rightarrow m(z) \ba^\top \bb$$ This comes directly from the deterministic equivalent of $\Q_0$
	\item $$\frac{1}{\sqrt{n}}\ba^\top \Q_0(z) \X \bb \rightarrow 0$$ This comes from the fact that $\EE \frac{1}{\sqrt{n}}\Q_0 X = 0$ and a concentration of measure argument
	\item $$\frac{1}{n} \ba^\top \X^\top \Q_0(z) \X \bb \rightarrow (zcm(z) + c)\ba^\top \bb$$
	This comes from the identity $\frac 1n \X^\top \Q_0 \X = z(\frac 1n  \X^\top \X - z \I_n)^{-1} + \I_n$. For which we can recognise this as (almost) a resolvent of the transposed data matrix $\X^\top$.
	This has deterministic equivalent $(z \tilde{m}(z) + 1) \I_n$, where $\tilde{m}$ is the corresponding ``transposed'' Stieltjes transform. $\tilde{m}$ is a rescaling of $m$ with $c \mapsto c^{-1}$,
	 and then also correcting for the fact that we're dividing by $n$ instead of $p$.
	We may massage out the identity $\tilde{m}(z) = c m(z) - \frac{1-c}{z}$ to give the result
\end{enumerate}

These identities then give

$$\V^\top \Q_0 \U = 
\begin{pmatrix}
\frac{\tau}{\sqrt{n}} \bb^\top \X^\top \Q_0 \ba & \frac{1}{n} \bb^\top \X^\top \Q_0 \X \bb & \frac{\tau}{\sqrt{n}} \bb^\top \X^\top \Q_0 \ba \\
\tau^2 \ba^\top \Q_0 \ba & \frac{\tau}{\sqrt{n}} \ba^\top \Q_0 \X \bb & \tau^2 \ba^\top \Q_0 \ba \\
\tau^2 \ba^\top \Q_0 \ba & \frac{\tau}{\sqrt{n}} \ba^\top \Q_0 \X \bb & \tau^2 \ba^\top \Q_0 \ba \\
\end{pmatrix}$$
$$\longrightarrow
\begin{pmatrix}
0 & zcm(z) + c & 0 \\
\tau^2 m(z) & 0 & \tau^2 m(z) \\
\tau^2 m(z) & 0 & \tau^2 m(z) \\
\end{pmatrix}
$$

and hence,

$$\bar{\A} =
\begin{pmatrix}
	1 & zcm(z) + c & 0 \\
	\tau^2 m(z) & 1 & \tau^2 m(z) \\
	\tau^2 m(z) & 0 & 1 + \tau^2 m(z) \\
\end{pmatrix}^{-1}
$$

Next we turn to eliminating the randomness in $\U$ and $\V$. We define the deterministic matrices:
$$\bar{\U} = 
\begin{bmatrix}
\tau \ba & 0 & \tau \ba	
\end{bmatrix}
$$

$$\bar{\V} = 
\begin{bmatrix}
0& \tau \ba & \tau \ba	
\end{bmatrix}
$$

and claim that 
$$\EE \left[\Q_0 \U \bar{\A} \V^\top \Q_0 \right] = \EE \left[\Q_0 \bar{\U} \bar{\A} \bar{\V}^\top \Q_0 \right]$$

To see this, note firstly that $\U - \bar{\U}$ and $\V - \bar{\V}$ only contain terms of the form $\frac{1}{\sqrt n} \X \bb$.
If this term appears on its own in the expansion, then the expectation will be 0 since the matrix will then be an odd function of 
the centered data matrix $\X$.
It then only remains to show that the cross term $\frac1n \EE \left[ \Q_0 \X \bb \bb^\top \X^\top \Q_0  \right]$ is of vanishing size.
$$ \frac1n \EE \left[ \Q_0 \X \bb \bb^\top \X^\top \Q_0  \right] = \frac1n \sum_{i,j = 1}^n \EE \left[b_i b_j \Q_0 \x_i \x_j^\top \Q_0 \right]$$
$$ = \frac1n \sum_{i=1}^n \EE \left[b_i^2 \Q_0 \x_i \x_i^\top \Q_0 \right]$$
$$ =\frac1n  \sum_{i=1}^n \EE \left[ b_i^2 \frac{\Q_0^{-i} \x_i \x_i^\top \Q_0^{-i}}{(1 + \x_i^\top \Q_0^{-i} \x_i)^2} \right]$$

And then since each term is positive semidefinite, we may conclude:

$$ \left\| \frac1n \sum_{i=1}^n \EE \left[ b_i^2 \frac{\Q_0^{-i} \x_i \x_i^\top \Q_0^{-i}}{(1 + \x_i^\top \Q_0^{-i} \x_i)^2} \right] \right\| 
\leq  \left\| \frac1n \sum_{i=1}^n \EE \left[ b_i^2 \Q_0^{-i} \x_i \x_i^\top \Q_0^{-i} \right] \right\|$$
$$ = \frac1n \sum_{i=1}^n b_i^2 \left\|(\Q_0^{-i})^2\right\| = \|\bb\|^2 \left\|(\Q_0^{-i})^2\right\| / n = o(1)$$

Hence finally, we have shown that

$$\EE \left[ \Q_0 \U \A \V^\top \Q_0 \right] = \EE \left[ \Q_0 \bar{\U} \bar{\A} \bar{\V}^\top \Q_0 \right] + o_{\|\cdot\|}(1)$$
where now $\bar{\U} \bar{\A} \bar{\V}^\top$ is a purely deterministic matrix.

We may now argue that we can continue our deterministic replacement and replace the $\Q_0$ matrices with their deterministic equivalents also.
In general it is not true for a deterministic matrix $\B$ that $\Q_0 \B \Q_0 \longleftrightarrow \bar{\Q}_0 \B \bar{\Q}_0$, however in \cite{couillet2022random} 2.9.5, it was shown that
this does hold true when the matrix $\B$ is of finite rank.
Fortunately here, we are in this case since $\bar{\U} \bar{\A} \bar{\V}^\top$ is of rank 3, and so

$$\EE \left[ \Q_0 \U \A \V^\top \Q_0 \right] = \bar{\Q}_0 \bar{\U} \bar{\A} \bar{\V}^\top \bar{\Q}_0 = o_{\|\cdot\|}(1)$$

To conclude, we first compute
$$\bar{\Q}_0 \bar{\U} \bar{\A} \bar{\V}^\top \bar{\Q}_0$$

$$= m(z) \I_p  
\begin{bmatrix}
\tau \ba & 0 & \tau \ba	
\end{bmatrix}
\begin{pmatrix}
	1 & zcm(z) + c & 0 \\
	\tau^2 m(z) & 1 & \tau^2 m(z) \\
	\tau^2 m(z) & 0 & 1 + \tau^2 m(z) \\
\end{pmatrix}^{-1}
\begin{bmatrix}
0 \\ \tau \ba^\top \\ \tau \ba^\top	
\end{bmatrix}
m(z) \I_p
$$
And after the algebra, and using the identity $z c m(z)^2 - (1 - c - z) m(z) + 1 = 0$,
we arrive at the result
$$ \bar{\Q}_0 \bar{\U} \bar{\A} \bar{\V}^\top \bar{\Q}_0 = m(z) \left( 1 - \frac{1}{1 + \tau^2 (1 + zm(z))}\right) \ba \ba^\top$$

And therefore 
$$\EE [\Q_1] = \bar{\Q}_0 + \bar{\Q}_0 \bar{\U} \bar{\A} \bar{\V}^\top \bar{\Q}_0 + o_{\|\cdot\|}(1)$$
$$=m(z) \left( \I_p - \ba \ba^\top \left( 1 - \frac{1}{1 + \tau^2 (1 + zm(z))}\right) \right)  + o_{\|\cdot\|}(1)$$

% \diegoinsert{Make remark precise}

\begin{remark}[On concentration of measure]
Now that we have established the expectation, showing concentration of measure can be done by standard arguments. We don't write this out in full here, as they are very similar to the arguments already done for the concentration of $\ba^\top \hat \bbeta$ in the proof of Proposition 4.4. To control the concentration of $\tr \A \Q_1$ and $\ba^\top \Q_1 \bb$, a martingale Burkholder argument can similarly be applied, allowing us to derive almost sure convergence of these functionals to their mean, by controlling the difference $\Q_1 - \Q_1^{-i}$. 

\end{remark}
% 	In the above proof, we technically only showed the expectation
% 	part of the Deterministic equivalent, i.e. that $\| \EE \Q_1 - \bar{\Q}_1\| \rightarrow 0$.
\end{proof}

\subsection{Proof of Lemma \ref{lemma:q1_squared_det_equiv}}
\begin{proof}
	We aim to compute a deterministic equivalent for $\Q_1^2$. There are tempting but wrong routes.
	First, one might claim $\Q_1^2 \longleftrightarrow (\bar{\Q}_1)^2$, which is false since $\EE[\Q_1^2] \neq (\EE[\Q_1])^2$.

	Next, since $\Q_1(z) = \left(\frac1n \Z \Z^\top - z \I\right)^{-1}$, we have $\frac{d}{dz}\Q_1 = \Q_1^2$, so one could guess $\Q_1^2 \longleftrightarrow \frac{d}{dz}\bar{\Q}_1$. This is also wrong because $\EE[\Q_1] = \bar{\Q}_1 + o_{\|\cdot\|}(1)$ does not ensure the remainder stays small after differentiation. For the unpoisoned resolvent $\Q_0$, \cite[2.9.5]{couillet2022random} gives $\Q_0^2 \longleftrightarrow m'(z)\I_p$.

	We use $\A^{-1} - \B^{-1} = \A^{-1}(\B - \A)\B^{-1}$, which converts differences of resolvents into differences of their inverses.

	We also use
	\[
	\Q_1 \longleftrightarrow \bar{\Q}_1
	= m(z)\Big(\I_p - \ba \ba^\top \Big(1 - \tfrac{1}{1+\tau^2(1+z m(z))}\Big)\Big),
	\]
	so by Sherman--Morrison,
	\[
	\bar{\Q}_1^{-1} = \frac{1}{m(z)}\big[\I_p + \tau^2(1+z m(z))\,\ba \ba^\top\big].
	\]
	Using $m(z)=\big(\frac{1}{1+c m(z)}-z\big)^{-1}$, this is
	\[
	\bar{\Q}_1^{-1}
	= \Big(\tfrac{1}{1+c m(z)}-z\Big)\I_p + \tfrac{\tau^2}{1+c m(z)}\,\ba \ba^\top .
	\]

	Now,
	\[
	\EE[\Q_1^2]
	= \EE[\Q_1 \bar{\Q}_1] + \EE[\Q_1(\Q_1-\bar{\Q}_1)]
	= \bar{\Q}_1^2 + \EE\!\left[\Q_1^2\big((\bar{\Q}_1)^{-1}-\Q_1^{-1}\big)\right]\bar{\Q}_1
	\]
	\[
	= \bar{\Q}_1^2 + \EE\!\left[\Q_1^2\!\left(\tfrac{1}{1+c m(z)}\I_p - \tfrac1n \Z \Z^\top + \tfrac{\tau^2}{1+c m(z)}\,\ba \ba^\top\right)\right]\bar{\Q}_1
	\]
	\[
	= \bar{\Q}_1^2 + \frac{\EE[\Q_1^2]\bar{\Q}_1}{1+c m(z)}
	- \frac1n\sum_{i=1}^n \EE\!\left[\Q_1^2 \z_i \z_i^\top\right]\bar{\Q}_1
	+ \frac{\tau^2}{1+c m(z)}\,\EE[\Q_1^2]\,\ba \ba^\top \bar{\Q}_1 .
	\]

	For the middle term, use Sherman--Morrison to isolate the contribution of $\z_i$ in $\Q_1$ and then add and subtract $\bar{\Q}_1$:
	\begin{align*}
	\frac1n\sum_{i=1}^n \EE\!\left[\Q_1^2 \z_i \z_i^\top\right]\bar{\Q}_1
	&= \frac1n\sum_{i=1}^n \EE\!\left[\frac{\Q_1 \Q_1^{-i}\,\z_i \z_i^\top \bar{\Q}_1}{1+\frac1n \z_i^\top \Q_1^{-i}\z_i}\right] \\
	&= \frac1n\sum_{i=1}^n \EE\!\left[\frac{(\Q_1^{-i})^2 \z_i \z_i^\top \bar{\Q}_1}{1+\frac1n \z_i^\top \Q_1^{-i}\z_i}\right]
	- \frac1n\sum_{i=1}^n \EE\!\left[\frac{\frac1n \Q_1^{-i}\z_i \z_i^\top (\Q_1^{-i})^2 \z_i \z_i^\top \bar{\Q}_1}{\big(1+\frac1n \z_i^\top \Q_1^{-i}\z_i\big)^2}\right].
	\end{align*}
	Using the Quadratic Form Close-to-Trace lemma,
	\begin{align*}
	\frac1n\sum_{i=1}^n \EE\!\left[\Q_1^2 \z_i \z_i^\top\right]\bar{\Q}_1
	&= \frac1n\sum_{i=1}^n \EE\!\left[\frac{(\Q_1^{-i})^2 \z_i \z_i^\top \bar{\Q}_1}{1+c m(z)}\right]
	- \frac1n\sum_{i=1}^n \EE\!\left[\Big(\frac1n \tr (\Q_1^{-i})^2\Big)\frac{\Q_1^{-i}\z_i \z_i^\top \bar{\Q}_1}{(1+c m(z))^2}\right]
	+ o_{\|\cdot\|}(1).
	\end{align*}
	Since $\EE[(\Q_1^{-i})^2] = \EE[\Q_1^2] + o_{\|\cdot\|}(1)$ and
	$\EE\big[\frac1n\sum_{i=1}^n \z_i \z_i^\top\big] = \I_p + \tau^2 \ba \ba^\top$, and finite-rank perturbations do not affect normalized traces,
	\[
	\frac1n \tr (\Q_1^{-i})^2 = \frac1n \tr \Q_0^2 + o(1) = c\, m'(z) + o(1).
	\]
	Therefore
	\begin{align*}
	\EE[\Q_1^2]
	&= \bar{\Q}_1^2 + \frac1n\sum_{i=1}^n \EE\!\left[\frac{c\,m'(z)}{(1+c m(z))^2}\,\Q_1^{-i}\z_i \z_i^\top \bar{\Q}_1\right] + o_{\|\cdot\|}(1) \\
	&= \bar{\Q}_1^2\!\left[\Big(1+\frac{c\,m'(z)}{(1+c m(z))^2}\Big)\I_p
	+ \frac{\tau^2 c\,m'(z)}{(1+c m(z))^2}\,\ba \ba^\top \right] \\
	&= \frac{1}{m^2(z)}\,\bar{\Q}_1^2 \left[m'(z)\I_p + \tau^2\big(m'(z)-m^2(z)\big)\ba \ba^\top \right] \\
	&= m'(z)\I_p
	+ \left(\frac{m'(z)(\tau^2+1)- m(z)^2 \tau^2}{\big(1+\tau^2(1+ z m(z))\big)^2} - m'(z)\right)\ba \ba^\top .
	\end{align*}
\end{proof}

\end{document}